\documentclass[10pt,twocolumn,letterpaper]{article}

\usepackage{iccv}
\usepackage{times}
\usepackage{epsfig}
\usepackage{graphicx}
\usepackage{amsmath}
\usepackage{amssymb}
 \usepackage{multirow}
 \usepackage{color}
 \usepackage[dvipsnames]{xcolor}
 \usepackage{subcaption}
 \usepackage{soul}
\usepackage{comment}

\usepackage[normalem]{ulem}

\usepackage{appendix}
\usepackage{enumitem}
\usepackage{threeparttable}

\usepackage{hhline}
\usepackage{graphicx}
 \usepackage{float}
\usepackage{siunitx}
\usepackage{xcolor}
\usepackage{booktabs,colortbl, array}
\usepackage{pgfplotstable}
\pgfplotsset{compat=1.8}

\definecolor{rulecolor}{RGB}{0,71,171}
\definecolor{tableheadcolor}{gray}{0.92}
\newcommand{\topline}{ %
        \arrayrulecolor{rulecolor}\specialrule{0.1em}{\abovetopsep}{0pt}%
        \arrayrulecolor{tableheadcolor}\specialrule{\belowrulesep}{0pt}{0pt}%
        \arrayrulecolor{rulecolor}}
\newcommand{\midtopline}{ %
        \arrayrulecolor{tableheadcolor}\specialrule{\aboverulesep}{0pt}{0pt}%
        \arrayrulecolor{rulecolor}\specialrule{\lightrulewidth}{0pt}{0pt}%
        \arrayrulecolor{white}\specialrule{\belowrulesep}{0pt}{0pt}%
        \arrayrulecolor{rulecolor}}
\newcommand{\bottomline}{ %
        \arrayrulecolor{white}\specialrule{\aboverulesep}{0pt}{0pt}%
        \arrayrulecolor{rulecolor} %
        \specialrule{\heavyrulewidth}{0pt}{\belowbottomsep}}%

\pgfplotstableset{normal/.style ={%
        header=true,
        string type,
        font=\addfontfeature{}\small,
        column type=l,
        every odd row/.style={
            before row=
        },
        every head row/.style={
            before row={\topline\rowcolor{tableheadcolor}},
            after row={\midtopline}
        },
        every last row/.style={
            after row=\bottomline
        },
        col sep=&,
        row sep=\\
    }
}

\usepackage[pagebackref=true,breaklinks=true,letterpaper=true,colorlinks,bookmarks=false]{hyperref}

\iccvfinalcopy %

\begin{document}
\def\SP{~~}

\title{\vspace{-1.2cm}BPKD: Boundary Privileged Knowledge Distillation \\ For Semantic Segmentation}

\author{
	\textbf{Liyang Liu} $^{1}$
	\SP
	\textbf{Zihan Wang} $^{2~3}$
	\SP 
	\textbf{Minh Hieu Phan} $^{1}$
	\SP
	\textbf{Bowen Zhang} $^{1}$
	\SP
 	\textbf{Jinchao Ge} $^{1}$
	\SP
	\textbf{Yifan Liu} $^{1}$\thanks{
Corresponding author, Email: yifan.liu04@adelaide.edu.au.
} 
	 \\ [0.285cm]
 $^1$ The University of Adelaide, Australia  ~ ~ ~ ~   $^2$ The University of Queensland, Australia \\ $^3$ CSIRO’s Data61, Australia
    % \\ [0.285cm]
        % {\tt\small \{akide.liu,vuminhhieu.phan,b.zhang,jinchao.ge,yifan.liu04\}@adelaide.edu.au,zihan.wang@uq.edu.au}
}

\maketitle

\begin{abstract}
Current knowledge distillation approaches in semantic segmentation tend to adopt a holistic approach that treats all spatial locations equally.
However, for dense prediction, students' predictions on edge regions are highly uncertain due to contextual information leakage, requiring higher spatial sensitivity knowledge than the body regions. 
To address this challenge, this paper proposes a novel approach called boundary-privileged knowledge distillation (BPKD). BPKD distills the knowledge of the teacher model's body and edges separately to the compact student model.
Specifically, we employ two distinct loss functions: (i) edge loss, which aims to distinguish between ambiguous classes at the pixel level in edge regions; (ii) body loss, which utilizes shape constraints and selectively attends to the inner-semantic regions.
Our experiments demonstrate that the proposed BPKD method provides extensive refinements and aggregation for edge and body regions. 
Additionally, the method achieves state-of-the-art distillation performance for semantic segmentation on three popular benchmark datasets, highlighting its effectiveness and generalization ability. BPKD shows consistent improvements across a diverse array of lightweight segmentation structures, including both CNNs and transformers, underscoring its architecture-agnostic adaptability. The code is available at \url{https://github.com/AkideLiu/BPKD}.
\vspace{-1mm}
\end{abstract}

\section{Introduction}

\begin{figure}[htbp]
    \begin{subfigure}{0.23\textwidth}
        \includegraphics[width=\linewidth]{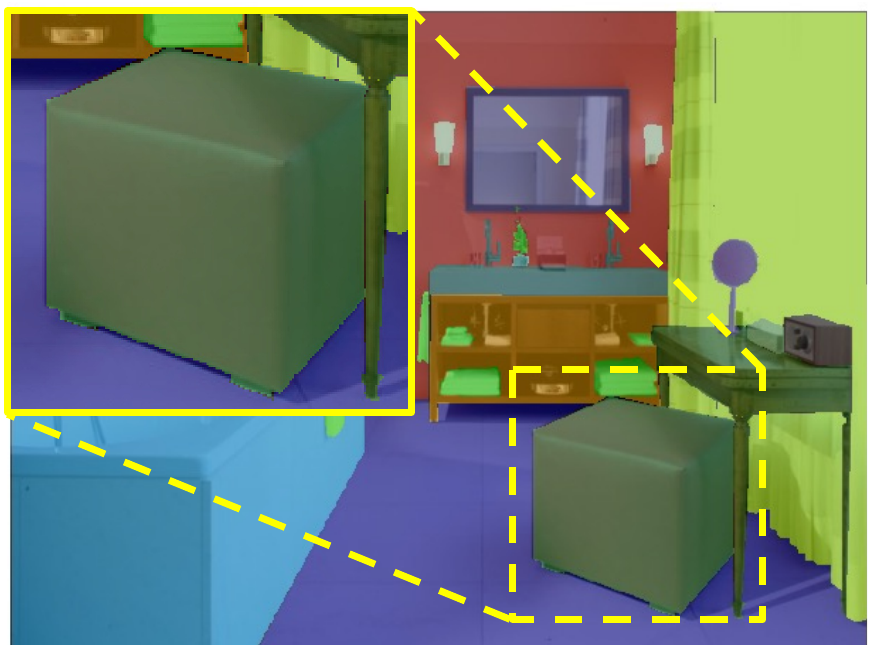}
        \caption{Original Image}
        \label{fig:sub1}
    \end{subfigure}
    \hfill
    \begin{subfigure}{0.23\textwidth}
        \includegraphics[width=\linewidth]{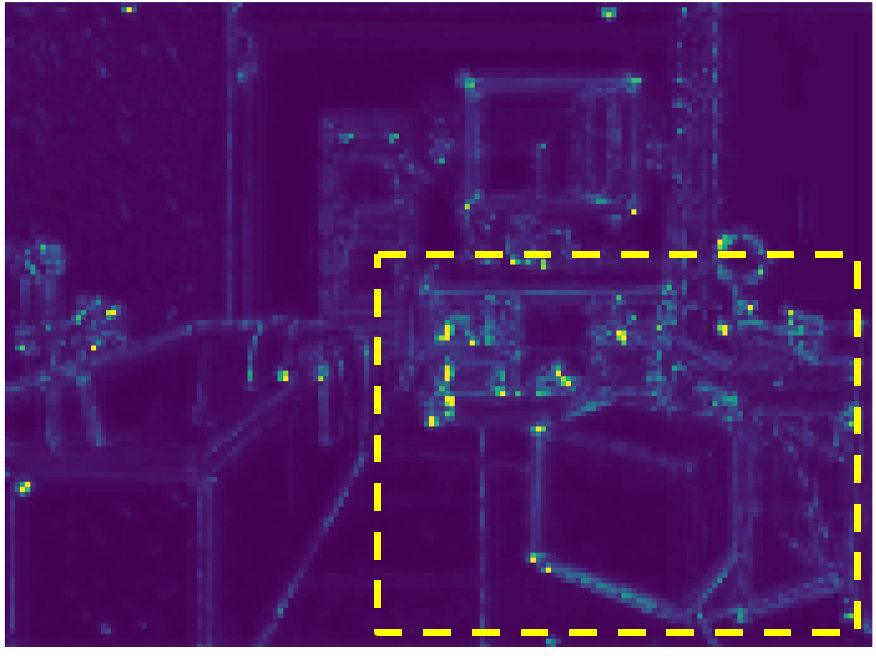}
        \caption{Student w/o KD}
        \label{fig:sub2}
    \end{subfigure}
    \\
    \begin{subfigure}{0.23\textwidth}
        \includegraphics[width=\linewidth]{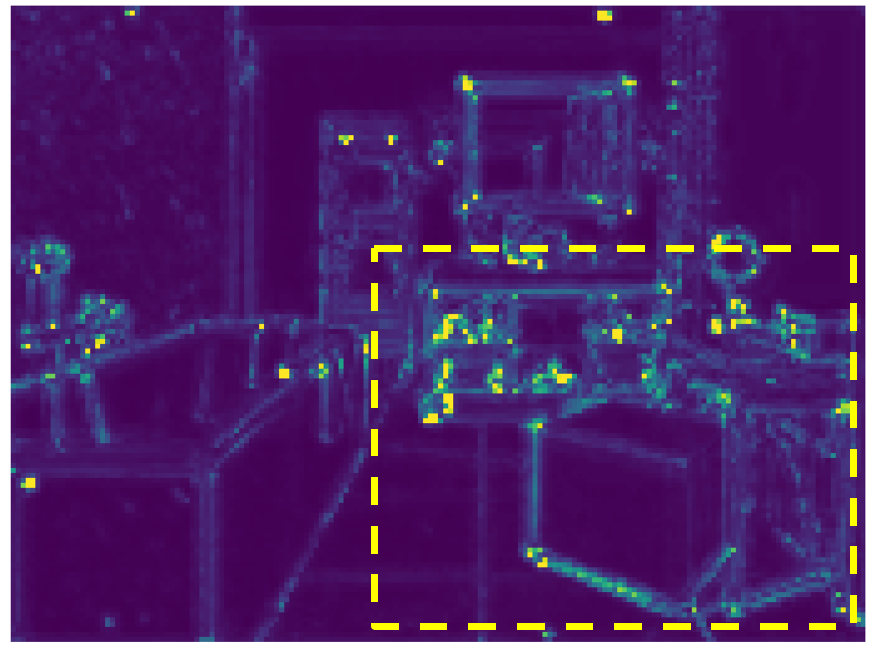}
        \caption{CWD}
        \label{fig:sub3}
    \end{subfigure}
    \hfill
    \begin{subfigure}{0.23\textwidth}
        \includegraphics[width=\linewidth]{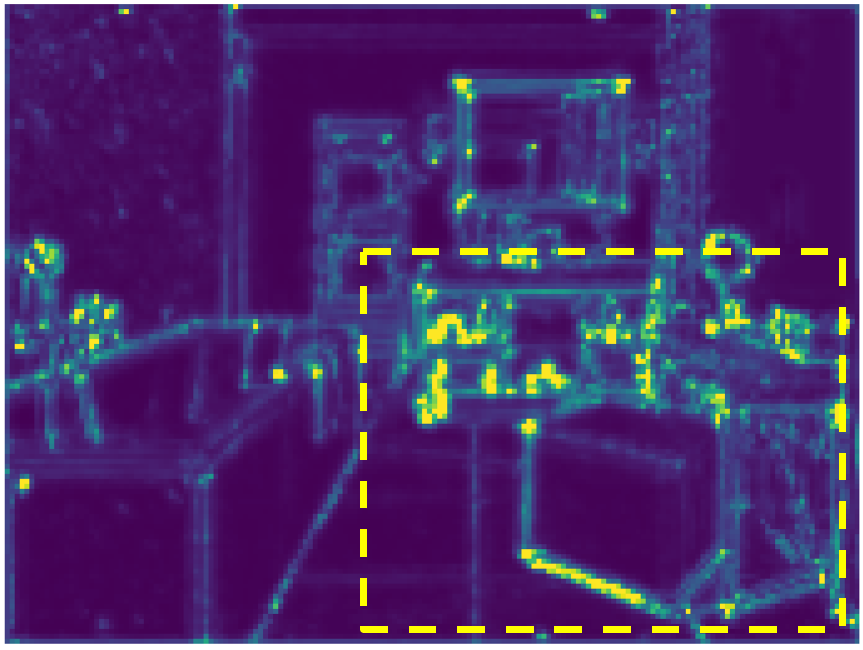}
        \caption{\textbf{BPKD(Ours)}}
        \label{fig:sub4}
    \end{subfigure}
    \scalebox{0.39}{\includegraphics{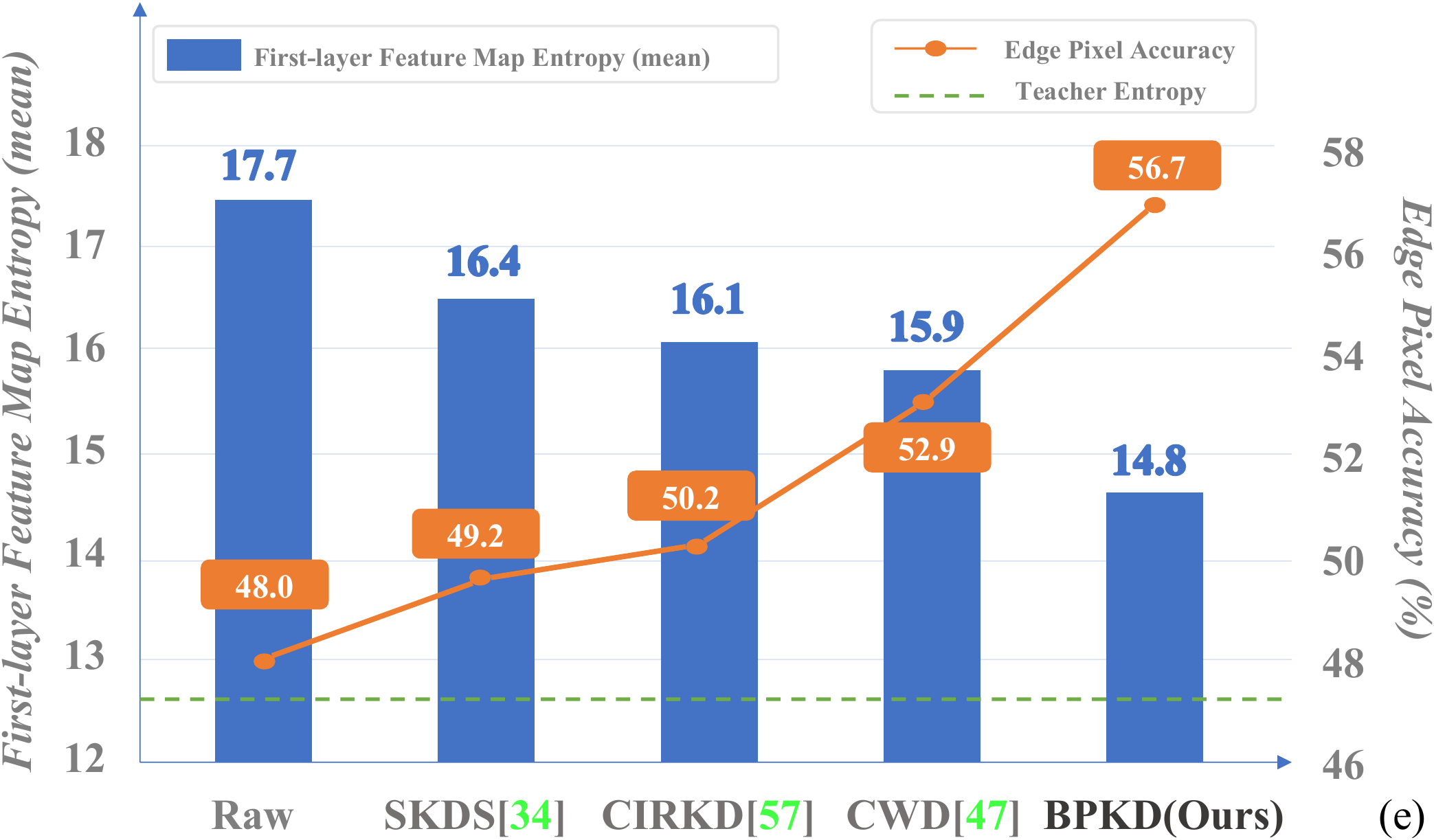}}
\caption{
 Illustration of contextual information leakage. Above) The uncertainty maps are generated by computing the mean entropy from first-layer feature maps, employing different distillation strategies such as the raw student, CWD, and BPKD. Brighter colors indicate higher certainty. Below) Demonstrates the inverse correlation between mean entropy and pixel-level accuracy along edges. When learning edge pixels, the model aggregates contextual information from adjacent class categories. Previous whole-view distillation methods are affected by contextual information leakage, leading to high uncertainty in edge low-level features and low edge pixel accuracy. 
 }
   \label{fig:Leaks}
     \vspace{-2em} %
\end{figure}

Semantic segmentation is a complex computer vision task that involves assigning unique categories to each pixel of an input image. In recent years, deep learning models with large numbers of parameters have achieved remarkable performance in semantic segmentation \cite{zhang2022resnest,liu2022convnet,he2016deep, zhang2022segvit, zhang2023segvitv2, cheng2021per,cheng2022masked}. However, such models are impractical for resource-constrained devices like mobile devices and robotics due to their high computational complexity \cite{wang2020deep,yuan2019segmentation,xie2021segformer}. To address this issue, lightweight base models such as MobileNet \cite{howard2019searching}, ShuffleNet \cite{ma2018shufflenet}, and EfficientNet \cite{tan2019efficientnet} have been used for real-time semantic segmentation.

Designing compression and acceleration techniques for compact networks is challenging but crucial. Knowledge distillation approaches, such as those introduced in \cite{hinton2015distilling,kim2018paraphrasing,zagoruyko2016paying,heo2019knowledge}, train a smaller student network to mimic the complex teacher network by minimizing the soft probabilities distance, typically measured by Kullback–Leibler (KL) divergence, between the student and teacher. In \cite{yim2017gift, lee2018self, zhang2018better}, authors have attempted to distill hidden knowledge by utilizing network and data relations, with a focus on classification tasks, achieving impressive results.

Pioneering knowledge distillation methods for semantic segmentation \cite{liu2019structured, shu2021channel, wang2020intra, yang2022cross} focus more on capturing the correlational information among pixels, channels, and images. Liu et al. \cite{liu2019structured} suggest that hidden knowledge in semantic segmentation is constructed by structured representation. Structured knowledge is more suitable for pair-wise similarity reduction and holistic distillation. IFVD\cite{wang2020intra} proposed to encode the knowledge according to the semantic masks. In CWD\cite{shu2021channel}, authors refine distillation by emphasizing aligning the most salient region of each channel between the teacher and student.

In comparison to prior studies, which encompassed a variety of studies including \cite{cao2022pkd, ji2022structural, liu2019structured, shu2021channel, wang2020intra, yang2022cross, yang2022masked}, that predominantly concentrated on transferring knowledge representations across the entire image,  the importance of distinct knowledge representations at different spatial locations has been neglected.
When learning edge features, the model aggregates contextual information between adjacent class categories, leading to \textit{contextual information leakage}. 
As shown in Fig.~\ref{fig:Leaks} b, c, d), current whole-view distillation methods exhibit high levels of uncertainty, as well as higher levels of entropy, at edge regions. Following prior works~\cite{abdar2021review}, we quantify the uncertainty by computing a mean entropy of first-layer features, capturing low-level textual representations. Fig.~\ref{fig:Leaks} e) shows that current methods suffer from higher uncertainties and lower accuracy at edge pixels, indicating the phenomenon of contextual information leakage. The low capacity of compact student networks further exacerbates this phenomenon, degrading segmentation details on the boundaries, especially for small object segmentation. However, delineating object's boundaries is mandatory for real-life applications such as localizing road boundaries for autonomous navigation~\cite{panda2023agronav} or segmenting tumors for treatment planning~\cite{liu2022self}.

To tackle the issue of contextual information learning in existing methods, we propose a novel approach, termed Boundary Privileged Knowledge Distillation (BPKD).
We divide the knowledge distillation process into two subsections: the edge distillation and the body distillation sections. 
Our proposed BPKD approach explicitly enhances the quality of edge regions and object boundaries by decoupling knowledge distillation and using teacher soft labels. The edge distillation loss involves spatial probability alignment and aggregation of contextual information to refine the boundaries. Furthermore, boundaries provide prior knowledge of the shape of an object's inner regions, and the body region can exploit this knowledge to eliminate high-uncertainty boundary samples and smooth the learning curves. Consequently, we observed that the object center received greater attention due to the implicit shape constraints, further improving segmentation in the body area.

Through empirical analysis, we have demonstrated that our proposed approach effectively guides the student network to learn from the teacher network's knowledge, resulting in improved segmentation performance. We evaluate our method over popular architectures on three segmentation benchmark datasets: Cityscapes\cite{cordts2016cityscapes}, ADE20K \cite{zhou2017scene}, and Pascal Context \cite{everingham2010pascal}. Experimental results indicate that BPKD outperforms other state-of-the-art distillation approaches. Specifically, we reduce the disparity in performance between the student and teacher networks and exhibit competitive results in comparison to specialized real-time segmentation methods\cite{fan2021rethinking}.

 Our main contributions are summarized as follows:

\begin{itemize}
  
  \item We show that current distillation methods suffer from contextual information leakage problems by analysing low-level feature uncertainty, leading to non-optimal segmentation performance at boundaries. To the best of our knowledge, this is the first paper identifying this critical problem within knowledge distillation literature for semantic segmentation.
  
  \item We propose a novel knowledge distillation method that separately focuses on distilling information related to the body and edge of objects. Our specialized edge loss function significantly enhances the quality of edge slices, while simultaneously imposing strong shape constraints on the body regions. This approach effectively minimizes the uncertainty prevents contextual information leakage in the distillation process and amplifies the focus on the inner region.
  
  \item Our method achieves state-of-the-art results on three popular benchmark datasets. We report an increase in the mean Intersection over Union (mIoU) by up to $4.02\%$ when compared to the previous SOTA CWD. Additionally, we observe a remarkable enhancement in prediction quality in both edge and body regions, further demonstrating our effectiveness.
\end{itemize}

\section{Related Work}

\label{sec:related}

\textbf{Semantic Segmentation.}  
Recent state-of-the-art approaches in semantic segmentation primarily leverage Fully Convolutional Networks (FCNs) \cite{liu2022cv,long2015fully,zhang2022segvit}. Notable models like PSPNet \cite{zhao2017pyramid} and DeepLab series \cite{chen2014semantic, chen2017deeplab, chen2017rethinking, chen2018encoder} employ advanced techniques such as pyramid pooling modules (PPM) and atrous spatial pyramid pooling (ASPP) to capture multi-scale contexts. HRNet \cite{wang2020deep} further innovates with a parallel backbone for high-resolution feature maintenance. Despite their performance, these models are computationally intensive, limiting their applicability in real-time and edge-device scenarios. Consequently, lightweight models like ENet \cite{paszke2016enet}, SqueezeNet \cite{iandola2016squeezenet}, and ESPNet \cite{mehta2018espnet} have gained traction. These models use strategies such as early downsampling, filter factorization, and efficient spatial pyramids to reduce computational overhead. MobileNet variants \cite{howard2017mobilenets,sandler2018mobilenetv2,howard2019searching} are also effective for efficient segmentation.

\textbf{Edge Detection.} Classical edge detection algorithms like Canny \cite{canny1986computational}, Sobel \cite{kanopoulos1988design}, and Prewitt \cite{prewitt1970object} have been retrofitted into modern deep learning architectures to achieve fine-grained segmentation. Deeply-supervised edge detection methods such as HED \cite{xie2015holistically} and RCF \cite{liu2019aquaculture} introduce multi-scale edge information directly into the segmentation pipelines. Similarly, models like CASENet \cite{yu2017casenet} have advanced the state-of-the-art by fusing class-specific edges into segmentation algorithms, providing a dual benefit of detailed boundary representation and class differentiation.
In parallel, techniques like edge-attention models \cite{wang2018non} incorporate edge information by weighting features based on their boundary importance, enabling finer contour mapping in semantic segmentation.

\textbf{Knowledge Distillation.} Knowledge distillation (KD) aims to condense the learnings from one or more expansive teacher models into a streamlined student model \cite{hinton2015distilling, gou2021knowledge}. Predominantly employed in basic vision tasks, KD techniques can be taxonomized into response-based, feature-based, and relation-based paradigms. Response-based methods, chiefly initiated by Hinton \textit{et al.} \cite{hinton2015distilling}, minimize Kullback-Leibler divergence to convey implicit, high-value knowledge \cite{zagoruyko2016paying, kim2018paraphrasing, heo2019knowledge}. Feature-based approaches like FitNet \cite{romero2014fitnets} align internal feature activations between teacher and student, whereas relation-based methods \cite{yim2017gift, lee2018self, zhang2018better} delve into inter-layer or inter-sample relationships. Nonetheless, traditional KD is largely skewed towards image classification, offering limited utility in pixel-level segmentation tasks.

Recent advancements have seen KD methods tailor-fit for semantic segmentation. Strategies such as structural knowledge distillation \cite{liu2019structured,liu2020structured} define segmentation as a structured prediction task, employing pair-wise similarities and holistic adversarial enhancements for knowledge transfer. Channel-wise distillation \cite{shu2021channel} concentrates on salient channel regions. Additional innovations include intra-class feature variation distillation \cite{wang2020intra}, which amalgamates pixel-level and class-wise variation, and Cross-Image Relational distillation \cite{yang2022cross}, which optimizes global semantic interconnections. Masked Generative Distillation \cite{yang2022masked} leverages teacher guidance for feature recovery. Recent work \cite{cao2022pkd} suggests Pearson correlation as a viable KL divergence alternative. Empirical validation corroborates the efficacy of these specialized KD techniques in boosting semantic segmentation performance.

In recent years, several techniques for knowledge distillation in semantic segmentation have been proposed, each focusing on different aspects of the problem. However, these methods have often neglected the significance of regional feature differences and their relative importance. In order to address this limitation, our approach involves splitting the knowledge distillation pipelines and improving the performance.

\section{Methods}

\begin{figure*}[t]
  \centering
    \scalebox{0.44}{\includegraphics{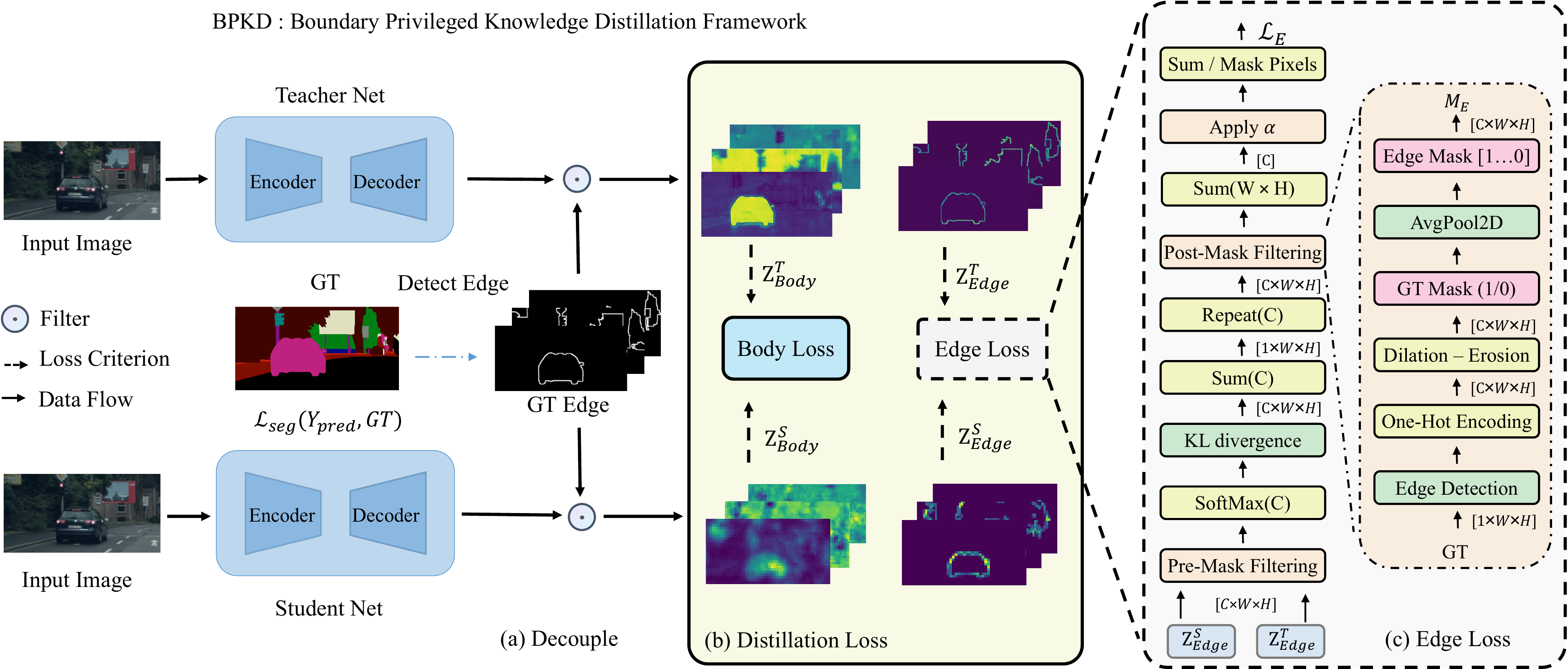}}
   \caption{
	Illustration of our proposed \textbf{Boundary Privileged Knowledge Distillation} framework and architecture. 
    (a) demonstrates the decoupling process that involves the edge detection on the ground truth to generate $GT_{Edge}$ Mask, followed by applying the mask filter to obtain the Teacher and Student logits masks. This step ensures that the information from the boundary region is isolated and appropriately conveyed to the Student.
    (b) shows that distillation comprises two terms: body loss and edge loss. The body loss term captures the categorized similarities, whereas the edge loss term concentrates on the boundary regions' transfer. $\mathcal{Z}_{E}$ and $\mathcal{Z}_{B}$ is short terms for $\mathcal{Z}_{Edge}$ and $\mathcal{Z}_{Body}$. In $\mathcal{Z}^{S, T}$, $S$ and $T$ standard for student and teacher, respectively. 
    (c) shows edge loss calculation is performed in two stages: pre-mask filtering and post-mask filtering. The pre-mask filtering step shapes the probability distribution to contain only edge information. Subsequently, the post-mask filtering step aggregates contextual information between adjacent categories to produce the final edge loss.
  	}
   \label{fig:architecture}
   \vspace{-3mm}
\end{figure*}

In this section, we first provide an overview of the workflow of the Boundary Privileged Knowledge Distillation (BPKD) framework (Section \ref{subsec:methods_ddrkd_framewok}), followed by a detailed description of two key implementation aspects of our approach. Specifically, Section \ref{subsec:edge_knowledge} outlines the edge knowledge distillation process, which involves pre-mask filtering and post-mask filtering. Section \ref{subsec:body_knowledge} introduces the distillation loss for body enhancements.

\subsection{BPKD Framework} 				\label{subsec:methods_ddrkd_framewok}

Existing feature distillation techniques~\cite{shu2021channel,yang2022cross} transfer the whole-view representations from the teacher while overlooking the effects of noisy edge features on the distillation process.
In this framework, we carefully consider the sensitivity of edge representations and introduce the novel boundary-privileged knowledge distillation (BPKD) that transfers the knowledge in the body and the edge regions separately. 
Distilling edge regions individually enhances the quality of object boundaries explicitly. Furthermore, the edge distillation loss provides prior shape knowledge for the object's inner regions. For instance, given a vehicle's boundary constraint, the model can easily determine pixel categories for its inner region. 
The body distillation loss has two key benefits from the prior boundary knowledge: (1) reducing learning difficulty by mimicking the teacher's logit probability distribution since high-uncertainty boundaries are removed, and (2) leveraging higher attention on the object center through implicit shape constraints.

Our approach uses an edge detection technique to generate edge masks $M_E$ for each class by processing the ground truth and the segmentation logit map. Let $\mathcal{Z} \in \mathbb{R}^{H \times W \times C}$ denote a network's logit map, where $C$ corresponds to the number of channels and $H \times W$ represents the spatial resolution. The edge masks $M_{E}$ are applied to separate the logit map $\mathcal{Z}$ into two components: the body component $\mathcal{Z}_{B}$ and the edge component $\mathcal{Z}_{E}$, which adhere to an additive rule, 
denoted by $\mathcal{Z} = \mathcal{Z}_{B} + \mathcal{Z}_{E}$.
Our BPKD framework separately transfers the edge and body knowledge encoded in these two components to the student.
As the edge slices have less amount of knowledge representations, we introduce a categorized awareness to balance the importance of different special perspectives. Together, these techniques play a crucial role in improving the overall performance of the model.

In this study, we propose a novel approach for decomposing the distillation loss into two distinct components, namely the body loss $\ell_{B}$ and the edge loss $\ell_{E}$, as expressed by Equation~\ref{math:bpkd1}. 
We include the body loss weight $\lambda_b$ and edge loss weight $\lambda_e$ to control the contributions of each loss term. This decoupling strategy allows us to examine the sensitivity of edge learning in the knowledge distillation process, which has been overlooked by the literature. The loss objective is defined as:
\\ 
\begin{equation} \label{math:bpkd1}
\begin{split}
  \ell= \lambda_b \cdot \ell_{B}\left(\mathcal{Z}_{B}^{S},\mathcal{Z}_{B}^{T}\right)+ \lambda_e \cdot \ell_{E}\left(\mathcal{Z}_{E}^{S}, \mathcal{Z}_{E}^{T}\right). 
\end{split}
\end{equation}

\subsection{Edge Knowledge Representation.~} \label{subsec:edge_knowledge}
Our framework minimizes the discrepancy between the teacher's and the student's features on the edge areas. To achieve this,
we extract an edge knowledge representation by using soft edge masks. This edge mask is applied on the logit map to produce a masked feature representation for teachers' and students' edge regions.
The edge map $M_E$ is created through two stages: Pre-Mask Filtering (PRM), which captures edge discrepancy for all classes, and Post-Mask Filtering (POM), which extracts edge discrepancy for each individual class. 
This approach allows a model to extract a more accurate and precise representation of the edge knowledge,
leading to improved performance in details classification.\\
During the edge detection process, we employ an adjustable Trimap algorithm \cite{wang2008image} to extract edge representations denoted by \(Z_{E}\) from the ground truth (GT). Although the teacher's predictions could serve as an alternative source for generating an edge mask, they offer slightly reduced accuracy compared to using GT directly. To generate the binary edge mask \(GT_{\text{edge}}\), we compute the difference between the dilation and erosion operations applied to the GT, formally expressed as \(GT_{\text{edge}} = \text{dilation}(GT) - \text{erosion}(GT)\). The resultant binary mask \(GT_{\text{edge}} \in \mathbb{R}^{C \times H \times W}\) undergoes average pooling to produce \(M_{E} \in \mathbb{R}^{C \times H' \times W'}\), which shares the same shape as the logits prediction \(\mathcal{Z}_{\text{pred}}\). The dimensions of \(M_{E}\) are governed by the output stride \(S\) of the segmentation network, specifically \(W' = W / S\).

\begin{itemize}
  \item 
  \textbf{Pre Mask Filtering (PRM)}. 
To obtain the logits map, we apply $M_{E}$ to both student and teacher logits: $\mathcal{Z}_{E} = \mathcal{Z}_{pred} \cdot M_{E}$. Specifically, we apply the edge mask for each channel $C$ so that we can concentrate the logits for the overlap edge regions. 
An intuitive example is if there is a frame that displays a dog and a cat standing nearby, only the logits activation of the dog and the cat class will be considered and all other activation will be suppressed. 
Such an operation forces the student to focus more on the correlations between the adjacent ambiguous classes. A spatial-level KL divergence loss is applied to the filtered logits $\mathcal{Z}_{E}^S$ and $\mathcal{Z}_{E}^T$:
\begin{equation} \label{math:bpkd1}
\begin{split}
\varphi(\mathcal{Z}_{E}^T,\mathcal{Z}_{E}^S) = \sum_{c=1}^C \phi(\mathcal{Z}_{E,i}^T) \cdot \log \frac{\phi(\mathcal{Z}_{E,i}^T)}{\phi(\mathcal{Z}_{E,i}^S)},
\end{split}
\end{equation}
where $\phi$ is the softmax operation for each pixel. $\varphi(\mathcal{Z}_{E}^T,\mathcal{Z}_{E}^S)$ represents the edge-masked KL distances for all spatial locations.

  \item 
  \textbf{Post Mask Filtering (POM)}. 
We further separate the edge loss for each class and perform normalization based on the edge area 
by Post Mask Filtering (POM). 
Let $\mathcal{Z}_{E,i,c}^T$ and $\mathcal{Z}_{E,i,c}^S$ denote the logits for the $c$-th class at pixel $i$ in the teacher and student models, respectively. Let $M_{E,c}$ be the soft mask obtained by average-pooling the ground truth binary edge mask for the $c$-th class, and $n_c$ denotes the number of non-zero pixels in $M_{E,c}$ for this class.  Our POM term can be formulated as follows:
\begin{equation} \label{math:edm_final}
\ell_{E} = \sum_{c=1}^C\frac{\alpha_c}{n_c} \sum_{i=1}^{W\cdot H} \varphi(\mathcal{Z}_{E,i,c}^T,\mathcal{Z}_{E,i,c}^S) \cdot M_{E,c},
\end{equation}

\noindent By re-weighing the loss based on the edge area of each class, we prioritize the center of the edge, where the most important information is often located.
This approach ensures that the student model focuses on learning the correct edge positions and shapes for each class.
\end{itemize}

\noindent The Soft Edge Masks $M_{E}$ play a critical role in the Edge Loss, and our approach to generating them involves two specialized designs: 1) converting binary $GT_{E}$ into a weighted discrete space, and 2) generating masks per channel instead of a unified mask. Directly applying binary masks may include unconfident bias, so we use average pooling to generate softer masks. We also carefully consider overlapping masks to minimize noise and uncertainty. Our mask design aims to exclusively include unconfident bias for minimizing knowledge distributions.

\noindent In summary, the proposed PRM and POM stages in the edge region refine the knowledge distillation process by identifying edge discrepancies for each class and applying a re-weighting to the loss based on the edge area. 
This method guarantees that the student model learns the correct edge positions and shapes for each class, and provides the shape prior knowledge for body knowledge representation.

\subsection{Body Knowledge Representation.~}\label{subsec:body_knowledge}
This section investigates the body knowledge distillation. Prior works consider whole-view distillation, which dilutes body knowledge with noisy representation on the edge.
To overcome these challenges, we utilize the reversed edge binary mask to extract body masks. By removing the edge region, we exploit implicit shape constraints and reduce uncertainty, which allows the body loss to focus on assigning the large inner regions of objects to their corresponding categories. To achieve this, we propose a region alignment approach that synthesizes channel-level activations to obtain semantically rich sections.
As we predefined that $\mathcal{Z} = \mathcal{Z}_{B} + \mathcal{Z}_{E}$. The body logits is obtained by $\mathcal{Z}_{B}=\mathcal{Z}\times (1-M_{E})$. As shown in the previous work~\cite{liu2019structured,shu2021channel}, a pixel-wise loss for the body region will bring in unexpected noise due to the hard constraints. Thus, we employ a loose constraint of the channel-wise distillation~\cite{shu2021channel} for the body part.
Body enhancement loss (BEL) is defined as:
\begin{equation}
\ell_{B}= \frac{\mathcal{T}^2}{C}\sum_{c = 1}^{C}\sum_{i=1}^{W\cdot H} 
	\phi (Z^{T}_{B,c,i}) \cdot \log \Bigl[
	\frac{\phi(Z^{T}_{B,c,i})}{\phi(Z^{S}_{B,c,i})}
	\Bigr],
	\label{eq:cw2}
\end{equation}

\newcommand{\tri}{\begin{small}\color{Green}$\blacktriangle$\end{small}}
\newcommand{\trid}{\begin{small}\color{BrickRed}$\blacktriangledown$\end{small}}
\newcommand{\trim}{\begin{small}\color{SpringGreen}$\blacktriangle$\end{small}}

\label{sec:table}
\begin{table*}
\centering
\caption{
Performance comparison of different distillation methods with state-of-the-art techniques. We test these methods on various segmentation networks for both student and teacher models, using datasets including Cityscapes\cite{cordts2016cityscapes}, ADE20K \cite{zhou2017scene}, and Pascal Context \cite{everingham2010pascal}. The FLOPs and FPS are obtained on $512\times512$ resolutions. \ Our BPKD outperforms all previous methods in large margins across multiple datasets and network architectures. DLab refers to Deeplab architecture. HRV2P refers to HRNetV2p. MV2 refers to MobileNet v2.
}
\resizebox{\linewidth}{!}{
\begin{tabular}{l|c|c|c|cc|cc|cc} 
\toprule
\multirow{2}{*}{}                    & \multirow{2}{*}{}          & \multirow{2}{*}{} & \multirow{2}{*}{}    & \multicolumn{2}{c|}{ADE20K}      & \multicolumn{2}{c|}{Cityscapes}   & \multicolumn{2}{c}{Pascal Context 59}  \\
Methods         & FLOPs(G) & Param(M) & FPS(S) & \multicolumn{2}{c|}{80k 512*512} & \multicolumn{2}{c|}{80k 1024*512} & \multicolumn{2}{c}{80k 480*480}        \\
                           &                   &         &          & mIoU(\%)       & mAcc(\%)        & mIoU(\%)            & mAcc(\%)         & mIoU(\%)            & mAcc(\%)              \\ 
\hline\hline
T: PSPNet-R101\cite{zhao2017pyramid}     & 256.89            & 68.07  & 2.68            & 44.39          & 54.75           & 79.74          & 86.56            & 52.47          & 63.15                 \\ 
\hline
S:PSPnet-R18\cite{zhao2017pyramid}      & 54.53             & 12.82   & 15.71          & 33.30          & 42.58           & 74.23          & 81.45            & 43.79          & 54.46                 \\
SKDS \cite{liu2019structured}                         & 54.53             & 12.82 & 15.71              & 34.49(\trim 1.19)          & 44.28           & 76.13(\trim1.9)          & 82.58            & 45.08(\trim1.29)          & 55.56                 \\
IFVD \cite{hou2020inter}                         & 54.53             & 12.82 & 15.71              & 34.54(\trim1.24)          & 44.26           & 75.35(\trim1.12)          & 82.86            & 45.97(\trim2.18)          & 56.6                  \\
CIRKD \cite{yang2022cross}                        & 54.53             & 12.82 & 15.71              & 35.07(\trim1.77)          & 45.38           & 76.03(\trim1.80)          & 82.56            & 45.62(\trim1.83)          & 56.15                 \\
CWD  \cite{shu2021channel}                          & 54.53             & 12.82 & 15.71              & 37.02(\trim3.72)          & 46.33           & 76.26(\trim2.03)          & 83.04             & 45.99(\trim2.20)          & 55.56                 \\
\textbf{BPKD(Ours)}              & 54.53             & 12.82 & 15.71              & \textbf{38.51(\tri 5.21)} & \textbf{47.70}  & \textbf{77.57(\tri 3.34)} & \textbf{84.47}   & \textbf{46.82(\tri 3.03)} & \textbf{56.29}                      \\ 
\hline\hline
T:HRV2P-W48 \cite{wang2020deep}    & 95.64             & 65.95 &6.42            & 42.02          & 53.52           & 80.65          & 87.39            & 51.12          & 61.39                 \\ 
\hline
S:HRV2P-W18S  \cite{wang2020deep}   & 10.49             & 3.97 & 23.74              & 31.38          & 41.39           & 75.31          & 83.71            & 40.62          & 51.43                 \\
SKDS  \cite{liu2019structured}                      & 10.49             & 3.97 & 23.74              & 32.57(\trim1.19)          & 43.22           & 77.27(\trim1.96)          & 84.77            & 41.54(\trim0.92)          & 52.18                 \\
IFVD  \cite{hou2020inter}                      & 10.49             & 3.97 & 23.74              & 32.66(\trim1.28)          & 43.23           & 77.18(\trim1.87)          & 84.74            & 41.55(\trim0.93)          & 52.24                 \\
CIRKD \cite{yang2022cross}                      & 10.49             & 3.97 & 23.74              & 33.06(\trim1.68)          & 44.30            & 77.36(\trim2.05)          & 84.97            & 42.02(\trim1.40)          & 52.88                 \\
CWD    \cite{shu2021channel}                      & 10.49             & 3.97 & 23.74              & 34.00(\trim2.62)          & 42.76           & 77.87(\trim2.56)          & 84.98            & 42.89(\trim2.27)          & 53.37                 \\
\textbf{BPKD(Ours)}              & 10.49             & 3.97 & 23.74              & \textbf{35.31(\tri 3.93)} & \textbf{46.11}  & \textbf{78.58(\tri 3.27)} & \textbf{85.78}   & \textbf{43.96(\tri 3.34)} & \textbf{54.51}        \\ 
\hline\hline
T:DLabV3P-R101 \cite{chen2017rethinking}  & 255.67            & 62.68  & 2.60           & 45.47          & 56.41           & 80.98          & 88.70             & 53.20          & 64.04                 \\ 
\hline
S:DLabV3P-MV2 \cite{sandler2018mobilenetv2}    & 69.60             & 15.35 & 8.40            & 31.56          & 45.14           & 75.29          & 83.11            & 41.01          & 52.92                 \\
SKDS   \cite{liu2019structured}                     & 69.60             & 15.35 & 8.40            & 32.49(\trim0.93)          & 46.47           & 76.05(\trim0.76)          & 84.14            & 42.07(\trim1.06)          & 55.06                 \\
IFVD  \cite{hou2020inter}                      & 69.60             & 15.35 & 8.40            & 32.11(\trim0.55)          & 46.07           & 76.97(\trim1.68)          & 84.85            & 41.73(\trim0.72)          & 54.34                 \\
CIRKD \cite{yang2022cross}                         & 69.60             & 15.35 & 8.40            & 32.24(\trim0.68)              & 46.09                 & 77.71(\trim2.42)          & 85.33            & 42.25(\trim1.24)              & 55.12                      \\
CWD  \cite{shu2021channel}                        & 69.60             & 15.35 & 8.40            & 35.12(\trim3.56)          & 49.76           & 77.97(\trim2.68)          & 86.68            & 43.74(\trim2.73)          & 56.37                 \\
\textbf{BPKD(Ours)}              & 69.60             & 15.35 & 8.40            & \textbf{35.49(\tri 3.93)} & \textbf{53.84}  & \textbf{78.59(\tri 3.30)} & \textbf{86.45}   & \textbf{46.23(\tri 5.22)} & \textbf{58.12}        \\ 
\hline\hline
T:ISANet-R101 \cite{huang2019isa}       & 228.21            & 56.80   & 2.35          & 43.80          & 54.39           & 80.61          & 88.29            & 53.41          & 64.04                 \\ 
\hline
S:ISANet-R18 \cite{huang2019isa}       & 54.33             & 12.46 & 17.34             & 31.15          & 41.21           & 73.62          & 80.36            & 44.05          & 54.67                 \\
SKDS  \cite{liu2019structured}                      & 54.33             & 12.46 & 17.34             & 32.16(\trim1.01)          & 41.80            & 74.99(\trim1.37)          & 82.61            & 45.69(\trim1.64)          & 56.27                 \\
IFVD  \cite{hou2020inter}                     & 54.33             & 12.46 & 17.34             & 32.78(\trim1.63)          & 42.61           & 75.35(\trim1.73)          & 82.86            & 46.75(\trim2.70)          & 56.4                  \\
CIRKD  \cite{yang2022cross}                     & 54.33             & 12.46 & 17.34             & 32.82(\trim1.67)          & 42.71           & 75.41(\trim1.79)          & 82.92            & 45.83(\trim1.78)          & 56.11                 \\
CWD    \cite{shu2021channel}                       & 54.33             & 12.46 & 17.34             & 37.56(\trim6.41)          & 45.79           & 75.43(\trim1.81)          & 82.64            & 46.76(\trim2.71)          & 56.48                 \\
\textbf{BPKD(Ours)}              & 54.33             & 12.46 & 17.34             & \textbf{38.73(\tri 7.58)} & \textbf{47.92}  & \textbf{75.72(\tri 2.10)} & \textbf{83.65}   & \textbf{47.25(\tri 3.20)} & \textbf{56.81}        \\
\hline
\end{tabular}}
\label{table:performance_city_pascal}
\vspace{-3mm}
\end{table*}

\section{Experiments}
\subsection{Experimental Setup}
\label{sec:intro}

\noindent\textbf{Dataset.} We conduct the experiments on three benchmark datasets for semantic segmentation: Cityscapes \cite{cordts2016cityscapes}, Pascal Context 2010 \cite{everingham2010pascal}, and ADE20K \cite{zhou2017scene}.

\noindent \textbf{ADE20K \cite{zhou2017scene}} contains 20k/2k/3k images for train/val/test with 150 semantic classes. It is constructed as the benchmark for scene parsing and instance segmentation.

\noindent \textbf{Cityscapes \cite{cordts2016cityscapes}} is an urban scene parsing dataset that contains 2975/500/1525 finely annotated images used for train/val/test. The performance is evaluated in 19 classes.

\noindent\textbf{Pascal Context \cite{everingham2010pascal}} provides dense annotations, which contain 4998/5105/9637 train/val/test images. We use 59 object categories for training and testing.  Our results are reported on the validation set.

\noindent\textbf{Implementation Details.} Our implementation is based on the open-source toolbox MMSegmentation \cite{mmseg2020,2021mmrazor} with PyTorch 1.11.0. We employ the standard data augmentation, including random flipping, cropping, and scaling in the range of [0.5, 2]. All experiments are optimized by SGD with a momentum of 0.9, and a batch size of 16. We use the crop of 512 $\times$ 512, 512 $\times$ 1024, and 480 $\times$ 480 for ADE20k, Cityscapes, and Pascal Context, correspondingly. We use an initial learning rate of 0.01 for ADE20K and Cityscapes. In addition, we use an initial learning rate of 0.004 for Pascal Context. The number of total training iterations is 80K. Following the previous methods \cite{chen2018encoder, zhao2017pyramid}, we use the poly learning rate policy 
and report the single-scale testing result. We conduct all experiments on 4 NVIDIA A100 GPUs. All the distillation methods are trained with the same configurations. 

\noindent\textbf{Metrics.} We set up a fair comparison by assigning identical parameters for each method with the same dataset. Mean Intersection-over-Union (mIoU), Trimap mIoU and pixel mean accuracy (mAcc) are employed as the main evaluation metrics. GFLOPs, FPS and No. Parameters are also reported for various student networks that we tested. All reported computational costs are measured using the fvcore.~\footnote{\url{https://github.com/facebookresearch/fvcore}} 

\begin{table}[!h]
\centering
\vspace{-0.5em}
\caption{Performance comparison of transformers-based architecture vs. different distillation strategies. Standard mIoU and Trimap mIoU of Swin Transformers \cite{liu2021swin} and DeiT \cite{touvron2021training} with ViT Adapter (DeiT-Ada) \cite{chen2022vision} on ADE20K with UPerNet \cite{xiao2018unified} decoder for 80K iterations. Distillation forward speed (DFS.), training time (TT.) and GPU memory footprint (GMem.) light our method have negligible computational cost. (DFS.) and (TT.) estimated on DeiT-Adapter with batch size = 16 with 4 GPUs and (GMem.) standard for per sample video memory allocation. 
}
\resizebox{\columnwidth}{!}{
\LARGE
\begin{tabular}{l|ccc|cc|cc} 
\toprule
 & \multicolumn{1}{l}{DFS.(S)↑} & \multicolumn{1}{l}{TT.(H)↓} & \multicolumn{1}{l|}{GMem.(G)↓} & \multicolumn{1}{l}{Swin↑} & \multicolumn{1}{l|}{Trimap ↑} & \multicolumn{1}{l}{DeiT-Ada.↑} & \multicolumn{1}{l}{Trimap ↑} \\ 
\hline\hline
T:Base & 9.52 & 11.26 & 8.32 & 50.13 & 40.10 & 48.80 & 39.72 \\
S:Tiny & 12.80 & 8.44 & 3.87 & 43.57 & 32.78 & 41.10 & 32.15 \\ 
\midrule
SKDS & 8.72 & 11.36 & 4.45 & 43.58 & 33.04 & 41.90 & 32.25 \\
IFVD & 6.06 & 16.45 & 8.97 & 43.75 & 32.90 & 41.16 & 32.11 \\
CIRKD & 7.70 & 16.35 & 10.70 & 43.32 & 32.68 & 41.64 & 32.23 \\
CWD & 8.76 & 11.15 & 4.45 & 44.99 & 33.73 & 44.25 & 33.49 \\
\textbf{BPKD} & 7.84 & 13.49 & 5.49 & \textbf{46.13} & \textbf{38.11} & \textbf{45.25} & \textbf{37.05} \\
\hline
\end{tabular}
}
\vspace{-1.5em}
\label{fig:trans}
\end{table}

\begin{table*}[t]
\centering
\begin{tabular}{c|c|c|c|c|c|c} 
\toprule
\multicolumn{7}{c}{Method}                                                                                                                                                                                                                                  \\ 
\hline\hline
\multicolumn{4}{c|}{Teacher: PSP-ResNet101}                                                                                                       & \multicolumn{3}{c}{79.74\%}                                                                              \\
\multicolumn{4}{c|}{Student: PSP-ResNet18~\,~}                                                                                                        & \multicolumn{3}{c}{Standard: 68.99\%~ ~ Trimap: 55.34\%}                                                 \\ 
\hline
\multicolumn{4}{c|}{Channel Wise Distillation \cite{shu2021channel} }                                                                                                        & \multicolumn{3}{c}{Standard: 74.29\%~ ~ Trimap: 57.34\%}                                                 \\ 
\multicolumn{4}{c|}{Pixel Wise Distillation \cite{liu2019structured} }                                                                                                        & \multicolumn{3}{c}{Standard: 69.33\%~ ~ Trimap: 53.82\%}                                                 \\ 
\hline\hline
\multicolumn{1}{c}{}      & \multicolumn{1}{c}{} & \multicolumn{1}{c}{\textbf{Body(C)}}          & \multicolumn{1}{c}{Body(P)}                   & \multicolumn{1}{c}{\textbf{Edge(P)}}          & \multicolumn{1}{c}{Edge(C)}    & \textbf{Ours}                    \\ 
\hline
\multirow{2}{*}{mIoU(\%)} & Standard~  & 74.17~(\tri 5.18) & 72.70~(\trim 3.71) & 71.63 (\trim 2.64) & 66.83 (\trid 2.16) & \textbf{75.94 (\tri 6.95)}  \\ & Trimap~ & 56.20 (\trim 0.86) & 54.12~(\trid 1.22) & 61.37 (\tri 6.03) & 51.76~(\trid 3.58) & \textbf{62.91 (\tri 7.57)}  \\
\hline
\end{tabular}
\caption{The effectiveness of the decoupling whole-view knowledge representation. The results show knowledge representation for different spatial locations should be considered, separately. C and P denote channel-wise and pixel-wise knowledge distillation, respectively.
} 
\label{table:methods_first}
\vspace{-4mm}
\end{table*}

\subsection{Compare with State-of-the-arts Methods}

\noindent To ensure a fair comparison, we have re-implemented a number of previously proposed knowledge distillation methods, including those by~\cite{liu2019structured, wang2020intra, shu2021channel, yang2022cross}. Subsequently, we benchmarked our BPKD method against various compact networks, such as PSPNet with ResNet18 backbone~\cite{zhao2017pyramid}, HRNet-W18~\cite{wang2020deep}, Deeplab-V3+ with MobileNetV2 backbone~\cite{sandler2018mobilenetv2}, ISANet with ResNet18~\cite{huang2019isa}, Swin Transformers \cite{liu2021swin} with UPerNet \cite{xiao2018unified} and DeiT\cite{touvron2021training}-Adapter\cite{chen2022vision} with UPerNet \cite{xiao2018unified}.

\noindent \textbf{Performance.} Table \ref{table:performance_city_pascal} reports our method's performance on the ADE20K validation set, where the proposed BPKD achieves state-of-the-art (SOTA) performance across multiple student networks. The distillation process enhances the mean Intersection over Union (mIoU) for student networks by up to 24.33\%. Notably, BPKD consistently outperforms the current SOTA, CWD, by 3.87\% across all evaluated network architectures. Additional results on the Pascal Context validation set indicate an average performance increase of 2\% over SOTA methods. 
Further experiments presented in Table \ref{fig:trans} reveal the method's efficacy on popular Transformer architectures like Swin and DeiT, where it outperforms CWD by up to 2.53\%. These findings underscore the architecture-agnostic nature of our approach. 
In terms of computational efficiency, BPKD achieves a distillation forward speed up to $29.3\%$, training duration reduction of $21.2\%$ and $21.9\%$, and a memory consumption (GMem.) reduction of $94.8\%$ and $63.4\%$, compared to previous methods such as CIRKD and IFVD. Additionally, we register a $13\%$ improvement in Trimap metrics over CWD on the Swin architecture, emphasizing the method's cost-effectiveness with SOTA performance.

\subsection{Ablation Study}
In this section, we comprehensively evaluate our BPKD under different settings. All ablation experiments are conducted on the Cityscapes dataset with T: PSPNet-R101 and S: PSPNet-R18. To decrease computational costs, we adopt a streamlined training configuration, including crop size reduced to 512 $\times$ 512, and training schedule to 40k iterations. More experiment results are shown in the supplementary.

\noindent \textbf{Effectiveness of Decoupled Knowledge.}
To verify the effectiveness of the proposed knowledge distillation approach, we evaluate the segmentation performance in the edge region in Table \ref{table:methods_first}. We evaluated the Trimap mIoU metric~\cite{chen2018encoder} when using channel-wise and pixel-wise normalization for network distillation. Channel-wise normalization led to a decline in both standard mIoU by \(2.16\%\) and Trimap mIoU by \(3.58\%\), indicating its sensitivity in edge regions. Conversely, pixel-wise distillation enhanced Trimap mIoU by \(6.03\%\), benefiting from our specialized Edge loss design that refines boundary quality~\cite{kohli2009robust,krahenbuhl2011efficient}.
The Body loss function shows an increase in standard mIoU by \(5.18\%\) when uses channel-wise and \(3.71\%\) when spatial-wise, emphasizing its effectiveness for body regions. However, it had a limited impact on Trimap's performance.
In summary, our optimal approach, BPKD, achieved the best mIoU scores of \(75.94 (+6.95)\%\) and \(62.91 (+7.59)\%\) on standard and Trimap evaluations, respectively. This highlights BPKD's capability to refine semantic boundaries and body regions through pixel-level alignment and context aggregation.

\begin{table}[h!]
\centering

\resizebox{\columnwidth}{!}{
\begin{tabular}{l|c|c|c} 
\toprule
Method            & mIoU(\%)       & IMP.(\%)        & mAcc(\%)        \\ 
\hhline{====}
Teacher           & 79.74          & -              & 86.56           \\ 
\hline
ResNet18          & 68.99          & -              & 75.19           \\ 
\hline
+ PRM             & 70.37          & \tri1.98           & 76.95           \\
+ POM             & 71.63          & \tri2.64           & 78.47           \\
+ BEL             & 74.17          & \tri5.18           & 80.47           \\
+ PRM + BEL       & 74.81          & \tri5.92           & 81.52           \\
+ POM + BEL       & 74.62          & \tri5.63           & 80.98           \\
+ PRM + POM + BEL & \textbf{75.94} & \tri\textbf{6.95} & \textbf{82.62}  \\
\hline
\end{tabular}
}
\begin{tablenotes}
    \item[1] \scriptsize{\textbf{PRM}: Pre-Mask Filtering \qquad \textbf{BEL}: Body Loss \qquad \textbf{POM}: Post-Mask Filtering}
\end{tablenotes}
\caption{The different locations apply the mask in the proposed method. (IMP.) refers to the improvement achieved by the student network.}
\vspace{-5mm}
\label{table:filter_location}
\end{table}

\begin{figure*}[htbp!]
\begin{minipage}[b]{0.4\linewidth}
\centering
\begin{tabular}{c|c|c|l} 
\hline
Edge Width & mIoU           & IMP.      & mAcc   \\ 
\hline\hline
3          & 73.89          & \tri4.90          & 81.24  \\
5          & 74.26          & \tri5.27          & 82.00  \\
7          & \textbf{75.94} & \tri\textbf{6.95} & \textbf{82.62}  \\
10         & 74.70          & \tri5.71          & 82.13  \\
15         & 74.36          & \tri5.37          & 82.02  \\
\hline
\end{tabular}
\captionof{table}{Performance comparison for various different edge width with PSPNet-R18 on the Cityscapes validation set, for a fair comparison, we rerun the same setting 3 times and measure mean mIoU for evaluation.}
\label{table:edge_width}
\end{minipage}
\hfill
\begin{minipage}[b]{0.57\linewidth}
\centering
    \scalebox{0.45}{
\includegraphics{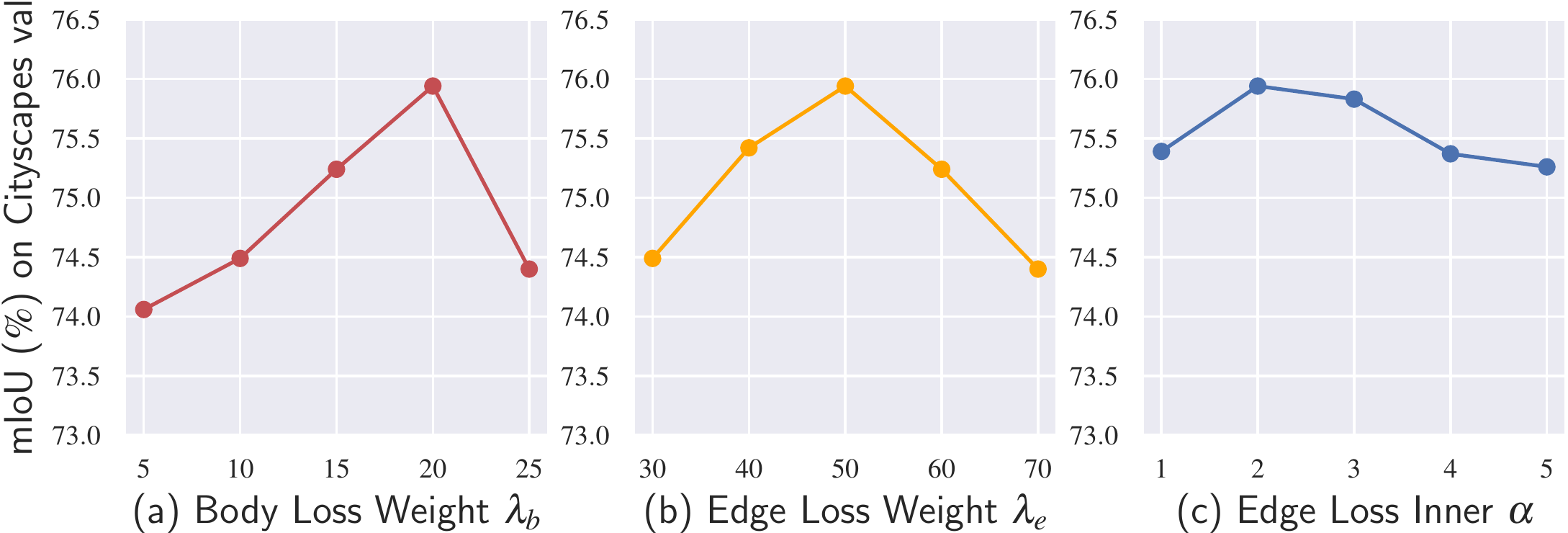}}
\captionof{figure}{Impact of the (a) Body Loss weight $\lambda_b$ and (b) Edge Loss weight $\lambda_e$ and (c) Edge Loss Inner weight $\alpha$ on Cityscape val. We found the optimal combination by board range study that $\lambda_b = 20, \lambda_e=50, \alpha=2$. 
}
\label{fig:hps}
\captionsetup{justification=centering}
\end{minipage}
\vspace{-5mm}
\end{figure*}

\begin{figure}[!b]
  \centering
    \scalebox{0.40}{
    \includegraphics{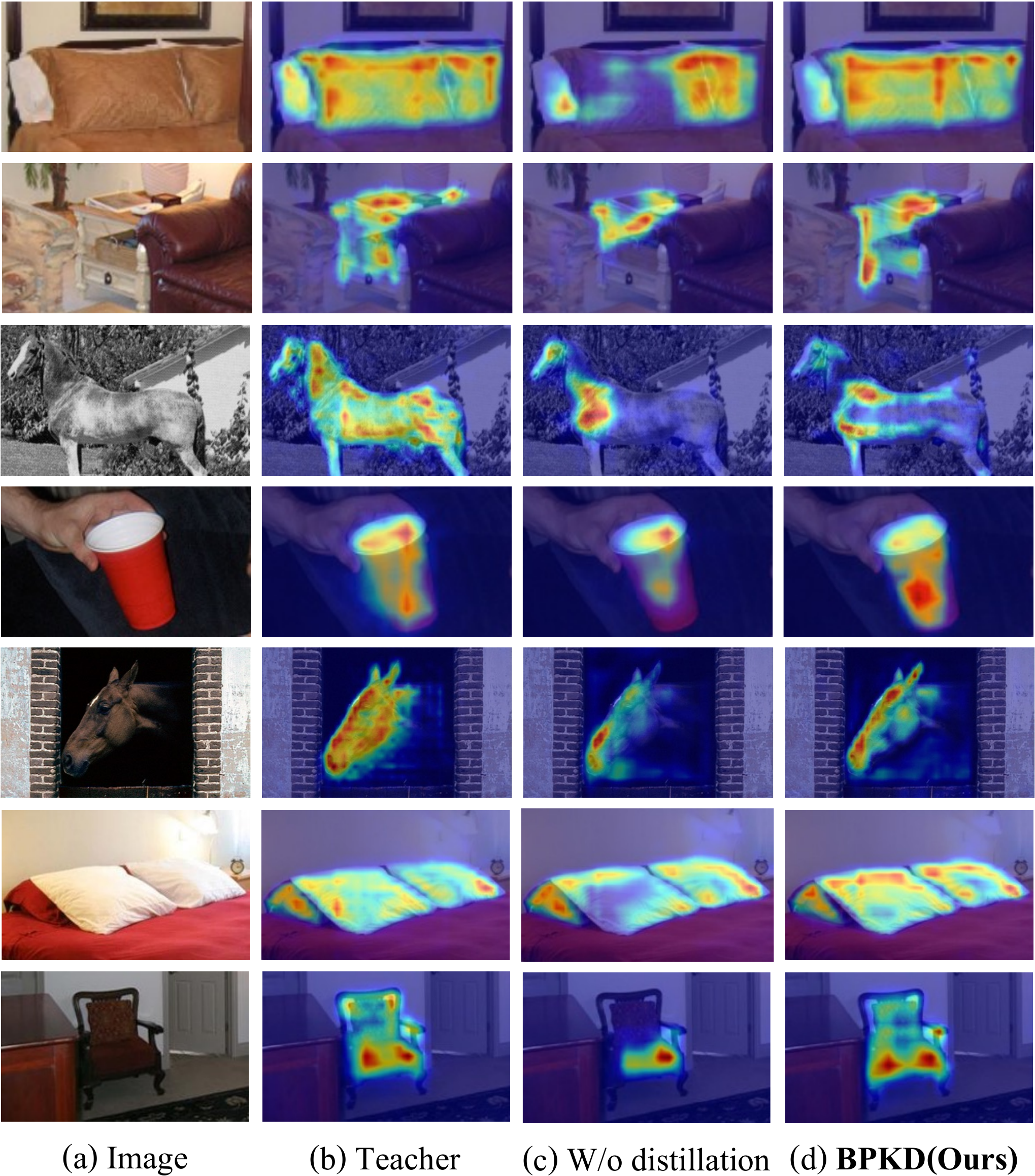}}
	\caption{
Comparison of CAM visualizations among (b) the teacher model, (c) the student model without distillation, and (d) the BPKD model. Activation maps were extracted from the last block of the corresponding ResNet backbones using HiResCAM \cite{draelos2020hirescam}. The results indicate that BPKD shows superior refinement of boundaries and higher attention to semantic bodies. For better visualization, zoom-in is recommended. %
}
	\label{fig:cam}
\end{figure}

\noindent \textbf{Compare Edge Filter Locations.} From Table \ref{table:filter_location}, the numerical results demonstrated the effectiveness of our proposed method. We further analyzed the impact of applying the edge filter for different locations. 
Applying the Pre Mask filter, the performance slightly improved by 1.98\% compared to the student without distillation. 
In contrast, Post Mask filtering improves the performance by 2.64\%, because POM extracts edge discrepancy specifically for each type of class.
The Body Enhancement Module takes care of non-edge information during our distillation setting, and the performance is raised by 5.18$\%$. 
Afterwards, we explore the performance by applying PRM or POM to BEL. 
The combination of three terms archives best mIoU that 75.94\% on the Cityscapes validation set.

\noindent \textbf{Impact of Different Edge Widths.} Edge width is pivotal in the Edge Detection Module, intricately influencing both the quality and the computational demands of the edge detection process. Specifically, a larger edge unit results in a wider edge, which incorporates more pixels into the edge loss. While this increase can potentially augment the robustness of edge detection, it also presents a trade-off: the wider the edge includes pixels from the body region, which could dilute the precision of edge-specific features.
We systematically analyze the impact of varying edge widths on the distillation performance, as summarized in Table \ref{table:edge_width}. Our analysis reveals that an edge width of 7 units is optimal, leading to a significant performance improvement of 6.95\%.

\noindent \textbf{Impact of Hyperparameters.} 
In the optimization process, the parameters \( \lambda_b \) and \( \lambda_e \) serve as weighting factors for the body and edge loss functions, respectively. These hyperparameters are pivotal in balancing the focus between semantic boundary refinement and overall body region accuracy. A higher value of \( \lambda_e \) places more emphasis on edge quality, conversely, a larger \( \lambda_b \) value makes the model more attuned to the larger body regions and potentially boosts the standard mIoU.  The edge loss incorporates an internal parameter \( \alpha \), serving as a class-wise balancing factor. This parameter is particularly instrumental in mitigating class imbalance by intra-adjusting the contribution of each class to the overall loss.  As shown in Fig \ref{fig:hps}, we investigate the impact of loss weights in our BPKD and find $\lambda_b=20$, $\lambda_e=50$ and $\alpha=2$ is the best choice.

\noindent \textbf{CAM Visualization Analysis.} Figure \ref{fig:cam} illustrates the explicit refinement of semantic boundaries on multiple objects. The shape constraints are evident, such as the strong attention given to the pillow outlines. Despite BPKD distillation, the student network cannot perfectly segment the horse in the second row, but it has highly attended to the bone silhouette. This shows that BPKD has tried its best to distill knowledge to the student network, but due to its limited size and capacity, the student can only learn the surface-level capacity of the teacher. The body sections have been smoothly affiliated to a single category, reducing high-uncertainty edge and incorporating shape prior knowledge from the edge loss pressure.
More qualitative segmentation results in supplement visually demonstrate our BPKD's effectiveness for both tiny and large objects with explicit boundaries enhancement.

\section{Conclusion}
This work presents a novel boundary-privileged knowledge distillation (BPKD) method for semantic segmentation, which transfers the cumbersome teacher model’s body and edge knowledge to the compact student model, separately. Extensive experiments demonstrate that the knowledge representation in the body and the edge regions should be considered differently. Due to different intrinsic properties, the edge region needs to focus on distinguishing between uncertain classes for each pixel while the body region needs to focus more on localizing and connecting object structures. Experimental results show that the proposed distillation method consistently outperforms state-of-the-art methods on various public benchmark datasets. The overall mIoU and the performance in the edge region are both improved by a large margin.

\section{Acknowledgments} 
Y. Liu acknowledges the support of start-up funding from The University of Adelaide for their participation in this work.
We express our gratitude to The University of Adelaide High-Performance Computing Services for providing the GPU Compute Resources, and to Mr. Wang Hui and Dr. Fabien Voisin for their valuable technical support.
We would like to thank Mr. Hanwen Wang for providing the visualization collection used in the supplementary.

% \newpage
% \newpage
{\small
\bibliographystyle{ieee_fullname}
\bibliography{main.bib}
}

\newpage

\begin{appendices}
\appendix

\begin{figure*}[ht]
  \centering
    \scalebox{0.35}{\includegraphics{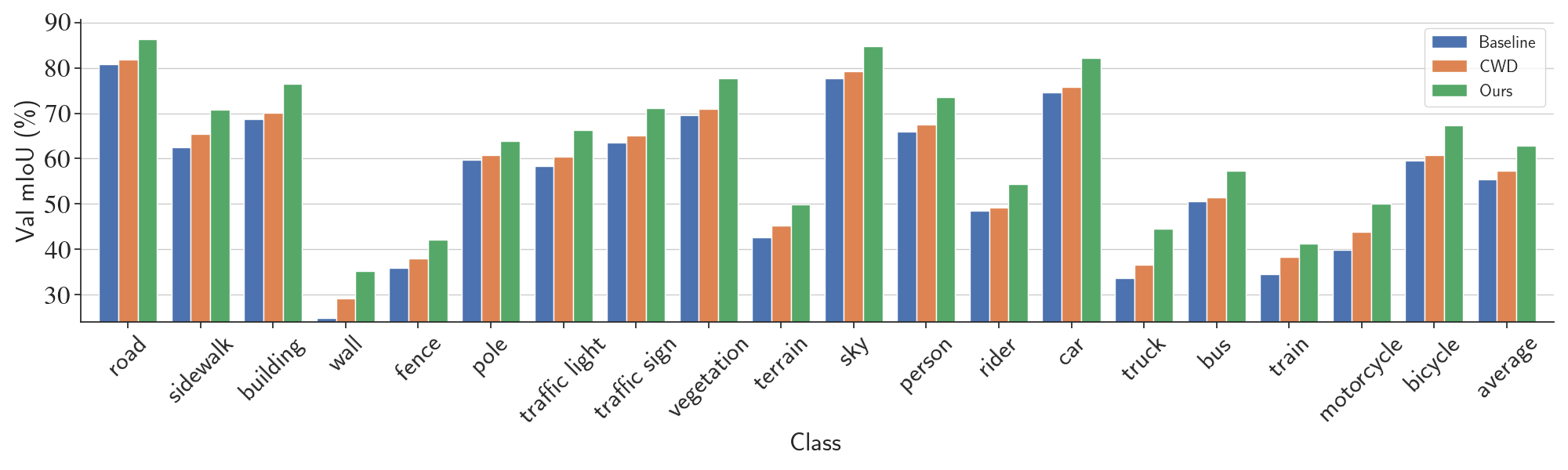}}
   \caption{
	Illustration of our proposed \textbf{Boundary Privilege Knowledge Distillation} schemes in terms of class \textbf{Trimap IoU} metrics with PSPnet + Resnet18 network architecture over the Cityscapes validation set. It can be seen from the figure that our method has different degrees of improvement for all categories, meanwhile , we have a significant improvement for categories that are difficult to distinguish by boundaries.
	}
   \label{fig:edge_iou}
      \vspace{-3mm}
\end{figure*}

In the supplementary materials, we augment the core arguments of our main paper with additional analyses, expanded experimental details, and enhanced visualizations. Initially, we conduct a performance comparison using reduced schedules across three unique datasets, specifically chosen for ablation studies. These comparative results are encapsulated in Table \ref{table:sota_supp}. To facilitate reproducibility, we provide an in-depth guide to our implementation, along with PyTorch-compatible code for our specialized Edge Loss. Additional analytical insights into class-trimap IoU are available in Figure \ref{fig:edge_iou} and Table \ref{table:trimap_iou}. Lastly, due to space constraints in the main manuscript, we include both visual and quantitative experiments on three distinct datasets within this supplementary section.

\begin{figure}[!bp]
  \centering
    \scalebox{0.49}{\includegraphics{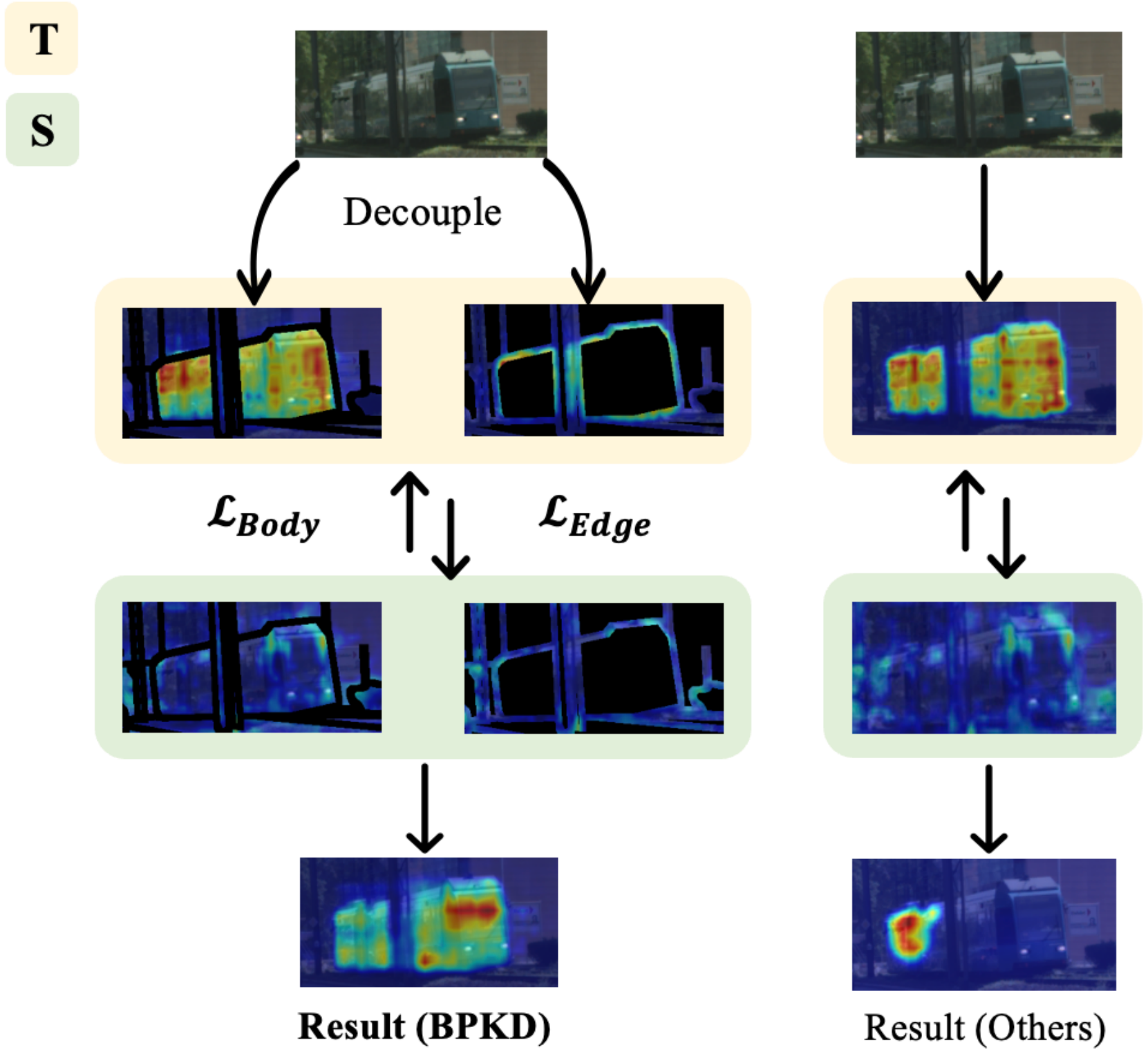}}
   \caption{
    BPKD\,(left) decouples the edge and body information from the input image and creates two parallel distillation pipelines compared to pioneering works, such as CWD \cite{shu2021channel}, SKDS \cite{liu2019structured},  IFVD \cite{wang2020intra}, CIRKD \cite{yang2022cross}. The proposed compositional loss forces the student to learn each part separately. The result shows the explicit performance gain. 
	}
   \label{fig:intro_diagram}
     \vspace{-2em} %
\end{figure}

\section{Performance On Reduced Schedules}
\noindent In order to demonstrate the efficacy of leveraging soft-labels from teachers to accelerate convergence speed, we conducted a comparison experiment with reduced schedules, 40k iteration training, and a $512\times512$ resolution. The results of this experiment, as well as comparisons with other state-of-the-art algorithms, are reported in Table \ref{table:sota_supp}. To ensure a fair comparison, our proposed knowledge distillation framework was applied to different teachers. Our students were able to learn knowledge from the teacher network, resulting in significant performance gains and achieving state-of-the-art results on multiple datasets. 
\noindent As demonstrated in Table \ref{table:sota_supp}, on ADE20K our method outperformed strong baseline channel-wise distillation by 2.34\%, 1.46\%, 1.66\%, 1.57\%, and 1.59\%, respectively. Furthermore, our methods were able to enhance the lightweight student network without increasing computational capacity, improving performance by 6.45\%, 4.13\%, 6.08\%, and 8.49\% compared to raw students.
\noindent  On Cityscape our method outperformed strong baseline channel-wise distillation by 2.22\%, 1.35\%, 4.40\%, 1.52\%, respectively. Furthermore, our methods were able to enhance the lightweight student network without increasing computational capacity, improving performance by 10.07\%, 3.78\%, 8.64\%, and 1.75\% compared to raw students.
 \noindent On Pascal Context our method outperformed strong baseline channel-wise distillation by 2.34\%, 1.46\%, 1.66\%, 1.57\%, and 1.59\%, respectively. Furthermore, our methods were able to enhance the lightweight student network without increasing computational capacity, improving performance by 10.07\%, 3.78\%, 6.08\%, and 8.49\% compared to raw students.
 \noindent  These numerical results suggest that our method is not dependent on a specific model structure, and that it produces significant performance gains with the pure student network, even without ImageNet Pre-train shows in table \ref{table:ade20k_no_imagenet}. To further demonstrate the effectiveness of our method, qualitative segmentation results are visualized in Figure \ref{fig:qualitative_ade}.

 \begin{table}[!htbp]
 \caption{Performance Comparison with state-of-the-art distillation methods on ADE20K dataset, the student backbone is not pre-trained on ImageNet.}
 \label{table:ade20k_no_imagenet}
\centering
\begin{tabular}{l|c|c} 
\toprule
Methods             & mIoU           & mAcc(\%)        \\ 
\midrule
T:PSPNet-R101 \cite{zhao2017pyramid}      & 44.39          & 54.74           \\ 
\hline
S:PSPnet-R18 \cite{zhao2017pyramid}       & 17.11          & 22.99           \\
SKDS  \cite{liu2019structured}                & 20.79          & 27.74           \\
IFVD  \cite{hou2020inter}               & 20.75          & 27.6            \\
CIRKD \cite{yang2022cross}              & 22.90          & 30.68           \\
CWD   \cite{shu2021channel}                & 24.79          & 31.44           \\
\textbf{BPKD(Ours)} & \textbf{27.46} & \textbf{36.10}  \\
\bottomrule
\end{tabular}
\end{table}

\begin{table*}[tbp]
\centering
\caption{
Performance comparison of different distillation methods with state-of-the-art techniques in \textbf{Reduced schedules}. We set a reduced training setting to reducethe computational cost, including crop size reduced to 512 × 512, and training schedule to 40k iterations. We test these methods on various segmentation networks for both student and teacher models, using datasets including Cityscapes\cite{cordts2016cityscapes}, ADE20K \cite{zhou2017scene}, and Pascal Context \cite{everingham2010pascal}. The FLOPs are obtained on $512\times512$ resolutions. \ Our BPKD outperforms all previous methods in large margins across multiple datasets and network architectures.
}
\label{table:sota_supp}
\resizebox{\linewidth}{!}{
\begin{tabular}{l|c|c|cc|cc|cc} 
\toprule
Methods & FLOPs(G) & Parameters(M) & \multicolumn{2}{c|}{ADE20K} & \multicolumn{2}{c|}{Cityscapes} & \multicolumn{2}{c}{Pascal Context 59} \\
 &  &  & \multicolumn{2}{c|}{40k 512*512} & \multicolumn{2}{c|}{40k 512*512} & \multicolumn{2}{c}{40k 512*512} \\
 &  &  & mIoU(\%) & mAcc(\%) & mIoU & mAcc(\%) & mIoU & mAcc(\%) \\ 
\hline\hline
T: PSPNet-R101 & 256.89 & 68.07 & 44.39 & 54.75 & 79.74 & 86.56 & 52.47 & 63.15 \\ 
\hline
S:PSPnet-R18 & 54.53 & 12.82 & 29.42 & 38.48 & 68.99 & 75.19 & 43.07 & 53.79 \\
SKDS & 54.53 & 12.82 & 31.80 & 42.25 & 69.33 & 75.37 & 43.93 & 54.01 \\
IFVD & 54.53 & 12.82 & 32.15 & 42.53 & 71.08 & 77.46 & 44.75 & 54.99 \\
CIRKD & 54.53 & 12.82 & 32.25 & 43.02 & 72.23 & 78.79 & 44.83 & 55.3 \\
CWD & 54.53 & 12.82 & 33.53 & 41.71 & 74.29 & 80.95 & 45.92 & 55.50 \\
\textbf{BPKD(Ours)} & 54.53 & 12.82 & \textbf{35.87} & \textbf{45.42} & \textbf{75.94} & \textbf{82.62} & \textbf{47.16} & \textbf{57.61} \\ 
\hline\hline
T: HRNetV2P-W48 & 95.64 & 65.95 & 42.02 & 53.52 & 80.65 & 87.39 & 51.12 & 61.39 \\ 
\hline
S:HRNetV2P-W18S & 10.49 & 3.97 & 28.69 & 37.86 & 73.77 & 82.89 & 40.82 & 51.70 \\
SKDS & 10.49 & 3.97 & 30.49 & 40.19 & 74.75 & 83.23 & 42.91 & 53.63 \\
IFVD & 10.49 & 3.97 & 30.57 & 40.42 & 75.33 & 83.83 & 43.12 & 54.03 \\
CIRKD & 10.49 & 3.97 & 31.34 & 41.45 & 74.63 & 83.72 & 43.45 & 54.10 \\
CWD & 10.49 & 3.97 & 31.36 & 40.24 & 75.54 & 84.08 & 45.50 & 56.01 \\
\textbf{BPKD(Ours)} & 10.49 & 3.97 & \textbf{32.82} & \textbf{43.49} & \textbf{76.56} & \textbf{85.34} & \textbf{46.12} & \textbf{57.63} \\ 
\hline\hline
T:DeeplabV3P-R101 & 255.67 & 62.68 & 45.47 & 56.41 & 80.98 & 88.7 & 53.2 & 64.04 \\ 
\hline
S:DeeplabV3P+MV2 & 69.6 & 15.35 & 22.38 & 31.71 & 70.49 & 80.11 & 37.16 & 49.1 \\
SKDS & 69.6 & 15.35 & 24.65 & 35.07 & 70.81 & 79.31 & 39.18 & 51.13 \\
IFVD & 69.6 & 15.35 & 24.53 & 35.13 & 71.82 & 80.88 & 38.8 & 50.79 \\
CIRKD & 69.6 & 15.35 & 25.21 & 36.17 & 72.39 & 81.84 & 39.99 & 52.66 \\
CWD & 69.6 & 15.35 & 26.89 & 35.79 & 73.35 & 82.41 & 42.52 & 53.24 \\
\textbf{BPKD(Ours)} & 69.6 & 15.35 & \textbf{28.46} & \textbf{41.45} & \textbf{76.58} & \textbf{84.14} & \textbf{44.32} & \textbf{56.04} \\ 
\hline\hline
T:ISANet-R101 & 228.21 & 56.8 & 43.8 & 54.39 & 80.61 & 88.29 & 52.94 & 63.52 \\ 
\hline
S: ISANet-R18 & 54.33 & 12.46 & 27.68 & 36.92 & 71.45 & 78.65 & 41.08 & 50.62 \\
SKDS & 54.33 & 12.46 & 28.70 & 38.51 & 70.65 & 77.53 & 42.87 & 52.89 \\
IFVD & 54.33 & 12.46 & 29.66 & 38.80 & 70.30 & 77.79 & 43.19 & 53.46 \\
CIRKD & 54.33 & 12.46 & 29.79 & 40.48 & 72.00 & 79.32 & 43.49 & 53.89 \\
CWD & 54.33 & 12.46 & 34.58 & 43.04 & 71.61 & 80.02 & 44.63 & 55.01 \\
\textbf{BPKD(Ours)} & 54.33 & 12.46 & \textbf{36.17} & \textbf{45.26} & \textbf{72.72} & \textbf{81.50} & \textbf{45.50} & \textbf{56.55} \\
\bottomrule
\end{tabular}
}
\vspace{-3mm}
\end{table*}

\begin{table*}[h]
\centering
\begin{tabular}{c|cccccccccc} 
\toprule
Class    & road  & sidewalk & building & wall  & fence & pole  & light & sign  & vege.   & terrain  \\ 
\hline\hline
Baseline                     & 80.88 & 62.44    & 68.72    & 24.82 & 35.91 & 59.73 & 58.39         & 63.58        & 69.64      & 42.58    \\
CWD                          & 81.84 & 65.36    & 70.14    & 29.22 & 37.98 & 60.72 & 60.43         & 65.12        & 70.91      & 45.28    \\
\textbf{BPKD(Ours)}          & 86.29 & 70.73    & 76.45    & 35.13 & 42.12 & 63.96 & 66.25         & 71.09        & 77.65      & 49.87    \\ 
\hline
Class    & sky   & person   & rider    & car   & truck & bus   & train & moto. & bicycle & average  \\ 
\hline\hline
Baseline                     & 77.72 & 65.94    & 48.45    & 74.55 & 33.63 & 50.58 & 34.58         & 39.83        & 59.56      & 55.34    \\
CWD                          & 79.2  & 67.58    & 49.11    & 75.76 & 36.53 & 51.37 & 38.29         & 43.82        & 60.75      & 57.34    \\
\textbf{\textbf{BPKD(Ours)}} & 84.87 & 73.56    & 54.4     & 82.28 & 44.49 & 57.27 & 41.27         & 50.14        & 67.37      & 62.91    \\
\bottomrule
\end{tabular}
\caption{Illustration of our proposed \textbf{Boundary Privilege Knowledge Distillation} schemes in terms of class \textbf{Trimap IoU} metrics with PSPnet + Resnet18 network architecture over the Cityscapes validation set.}
\label{table:trimap_iou}
      \vspace{-1mm}
\end{table*}

\section{Implementation Details of Edge Loss}

\noindent  This section presents the implementation details of the Edge Loss in PyTorch 
% shows  in \ref{fig:code}
, aimed at facilitating reproducibility. Our implementation consists of well-annotated components that are incorporated into our distillation system.
To begin with, the Pre Mask Filter operation is applied to the logits obtained from both teachers and students. Subsequently, a dimensional rearrangement is performed to optimize the process. The KL divergence serves as the core component to estimate the distance between the student and teacher probability distributions, which establishes an embryonic reference for calculating the loss. The Post Mask Filter takes input from the unreduced criterion and aggregates the distance on the channel dimension. It is followed by an additional spatial expansion that repeats the spatial information based on the given channels. Finally, the inner weights vector is applied to corresponding categories to enhance the learning capacity for hard edge samples. The EDM loss terms are then finalized by overall weights adaption and average reduction.
Furthermore, the Edge Detection Module is another crucial sub-component that employs multiple-level edge masks by providing input and ground truth labels. We utilize dilation and erosion to retrieve the edges, with the hyperparameter, $kernel\_size$, controlling the edge width. To address the computational pressure arising from a large kernel, the $compute\_iters$ is introduced for GPU memory optimization. Additionally, the edge detection module employs average pooling as a downsampling policy, considering progressively decreasing importance from internal boundaries to outlines.

\section{Categorical trimap Performance}

This section presents an evaluation of the categorical Edge IoU using the PSPnet encoder and Resnet18 backbone architecture on the Cityscape validation set. Figure \ref{fig:edge_iou} and Table \ref{table:trimap_iou} illustrate the results. Our proposed methods demonstrated significant improvement in most classes based on edge factors compared to the raw student model and strong baseline channel-wise distillation method. We also observed explicit improvement for categories that are difficult to distinguish by boundaries. For instance, we achieved a 10.31\% improvement for the wall category and a 6.69\% improvement for the bus category.

\begin{table}[!bp]
\centering
\resizebox{\columnwidth}{!}{
\begin{tabular}{l|cccccc} 
\hline
Methods       & $AP$   & $AP_{50}$ & $AP_{75}$ & $AP_{S}$  & $AP_{M}$  & $AP_{L}$   \\ 
\hline\hline
T:SOLOv2-X101 & 41.7 & 63.2 & 45.1 & 18.0 & 45.0 & 61.6  \\ 
\hline
S:SOLOv2-R18* & 26.7 & 44.1 & 27.5 & 6.50  & 27.2 & 45.8  \\
CWD           & 28.4 & 47.1 & 29.6 & 9.90  & 30.3 & 44.2  \\
\textbf{BPKD(Ours)}    & \textbf{33.2} &\textbf{53.6} & \textbf{35.0} & \textbf{13.4} & \textbf{36.0} & \textbf{50.6}  \\
\hline
\end{tabular}
}
\caption{The instance segmentation results presented in this report were obtained on the COCO\cite{lin2014microsoft} validation set using single-model results. All distillation methods and the student network baseline were trained using a 1x schedule with multiple-scale training disabled. The table below demonstrates that our distillation method can easily adapt to instance segmentation tasks and outperforms previous methods in a small-scale training setting. }
\label{fig:instance_seg}
\end{table}

\section{Instance segmentation}
\noindent We conducted experiments using SOLOv2 on the COCO dataset to demonstrate the general adaptability of our method. Specifically, we selected SOLOV2 \cite{wang2020solov2} X-101(DCN) as the teacher and Light SOLOV2 \cite{wang2020solov2} R-18 as the student. Table \ref{fig:instance_seg} presents the results, which show that our method improved the raw student by 6.5\%, 9.5\%, 7.5\%, 6.9\%, 8.8\%, and 4.8\% on the corresponding metrics.
\noindent The results demonstrate that our distillation method can easily adapt to instance segmentation tasks and outperforms previous methods in the small-scale training setting. While instance segmentation and semantic segmentation tasks have distinct differences, they share similar properties in that they predict masks for target senses and given pixel-level annotations. During the experiments, we applied knowledge distillation methods multiple times on pyramid classification logits and masked representations, followed by calculating the average across all levels of sub-terms to obtain the distillation loss.
\noindent From the numerical results, AP metrics explicitly increased for small and medium objects. However, there was no performance improvement for large objects, indicating that our method has room for improvement in instance segmentation tasks. As a future prospect, we aim to adjust the current method and design a specialized loss term for instance-level knowledge distillation.

\newpage

\begin{figure*}[!htp]
  \centering
    \scalebox{0.52}{\includegraphics{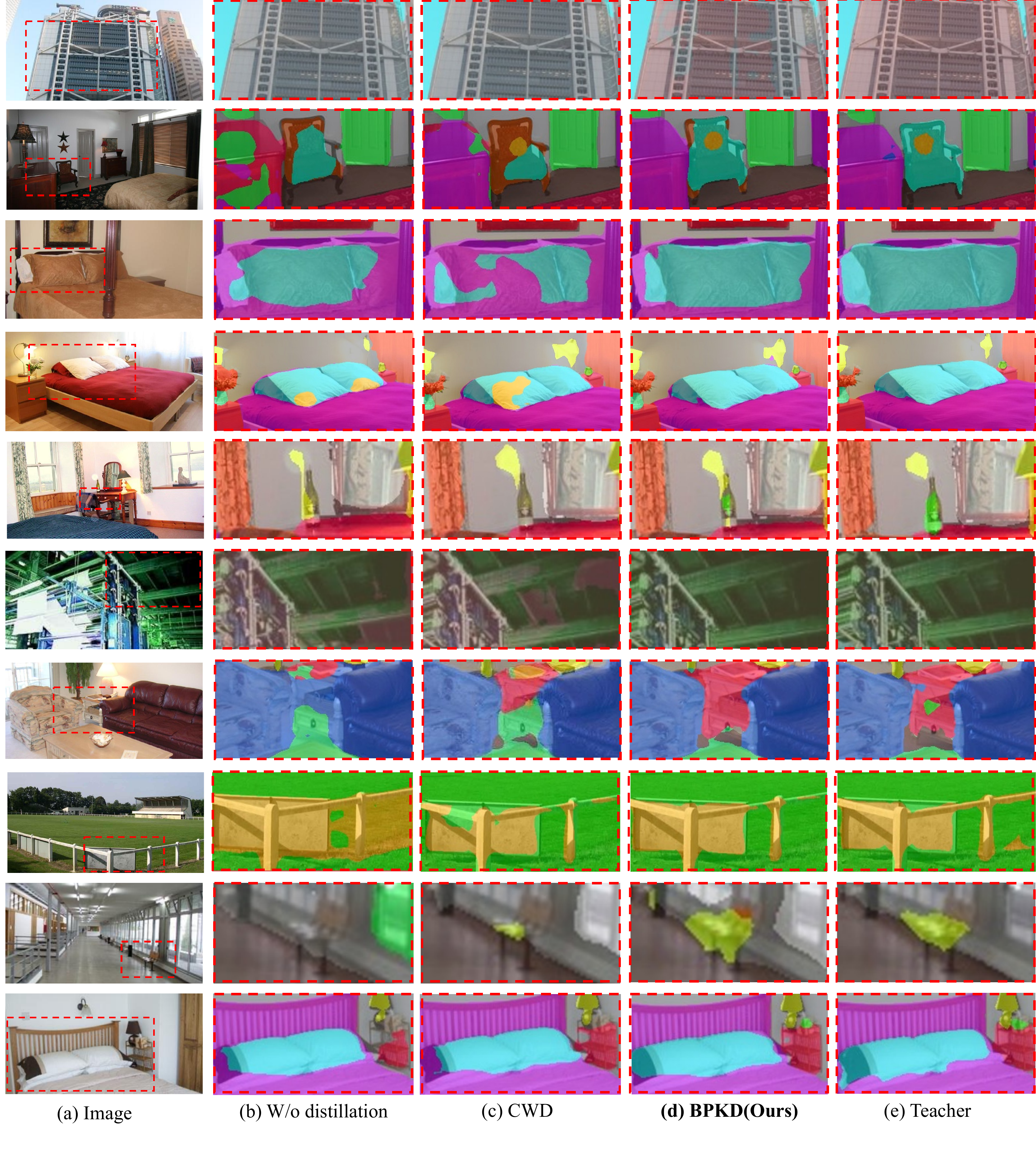}}
   \caption{
	Qualitative results on the ADE20K \cite{zhou2017scene} validation set produced by PSPNet \cite{zhao2017pyramid} and ResNet18 network architecture: (a) Initial images, (b) w/o distillation scheme, (c) sate-of-the-art method channel-wise distillation \cite{shu2021channel}, (d) \textbf{BPKD} our method, (e) teacher. This figure shows that our methods segment the small complex objects with explicit boundaries. Zoom in for a better view. 
 In the second row, BPKD demonstrates superior segmentation results; for instance, the contour of the chair is more distinct and the overall accuracy is enhanced. In the third, fourth, and last rows, the contours of the pillows are clearly delineated, coming closer to the results generated by the Teacher. In the seventh row, the coffee table situated between the two chairs is rendered with greater clarity, thereby suggesting that our method achieves commendable performance in resolving the ambiguity among multiple proximate objects. These visual outcomes indicate that our approach not only improves semantic boundaries but also leverages prior knowledge of contours and shapes to produce outstanding segmentation results.
	}
   \label{fig:qualitative_ade}
   \vspace{-1em}
\end{figure*}

\begin{figure*}[!htp]
  \centering
    \scalebox{0.52}{\includegraphics{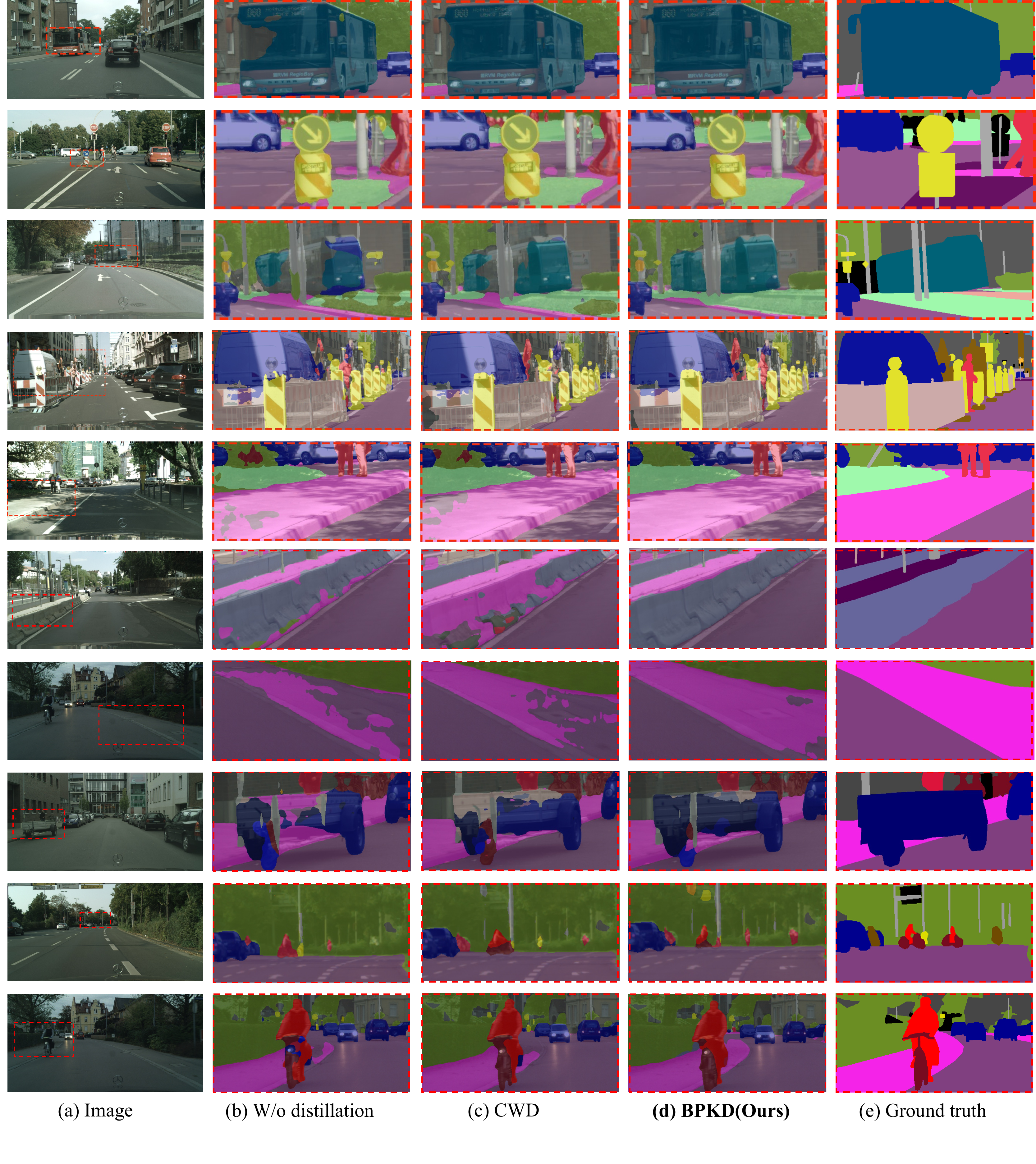}}
   \caption{
	Qualitative results on the Cityscapes \cite{cordts2016cityscapes} validation set produced by PSPNet \cite{zhao2017pyramid} and ResNet18 network architecture: (a) Initial images, (b) w/o distillation, (c) sate-of-the-art method channel-wise distillation \cite{shu2021channel}, (d) \textbf{BPKD} our method, (e) Ground truth. This figure shows that our methods segment the small complex objects with explicit boundaries. Zoom in for a better view.
 In the first row, BPKD effectively addresses the issue of the front windshield of the bus being wrongly segmented into multiple classes. In the second row, the approach enhances the contours of the upper and lower sections of the road signs. The third row presents a particularly striking result in the segmentation of a train; through differentiated supervision and distillation of the edges and main body, the train is clearly segregated from obstructions. In the fourth row, distant pedestrians present a challenging case for segmentation and identification due to their far-off camera angle and significant occlusions; despite these complexities, BPKD still manages to produce stable edges around the human figures. In the remaining images, BPKD consistently shows improvements, achieving commendable segmentation outcomes on both pedestrian paths and roadways.
	}
   \label{fig:qualitative_ade}
   \vspace{-1em}
\end{figure*}

\begin{figure*}[!htp]
  \centering
    \scalebox{0.52}{\includegraphics{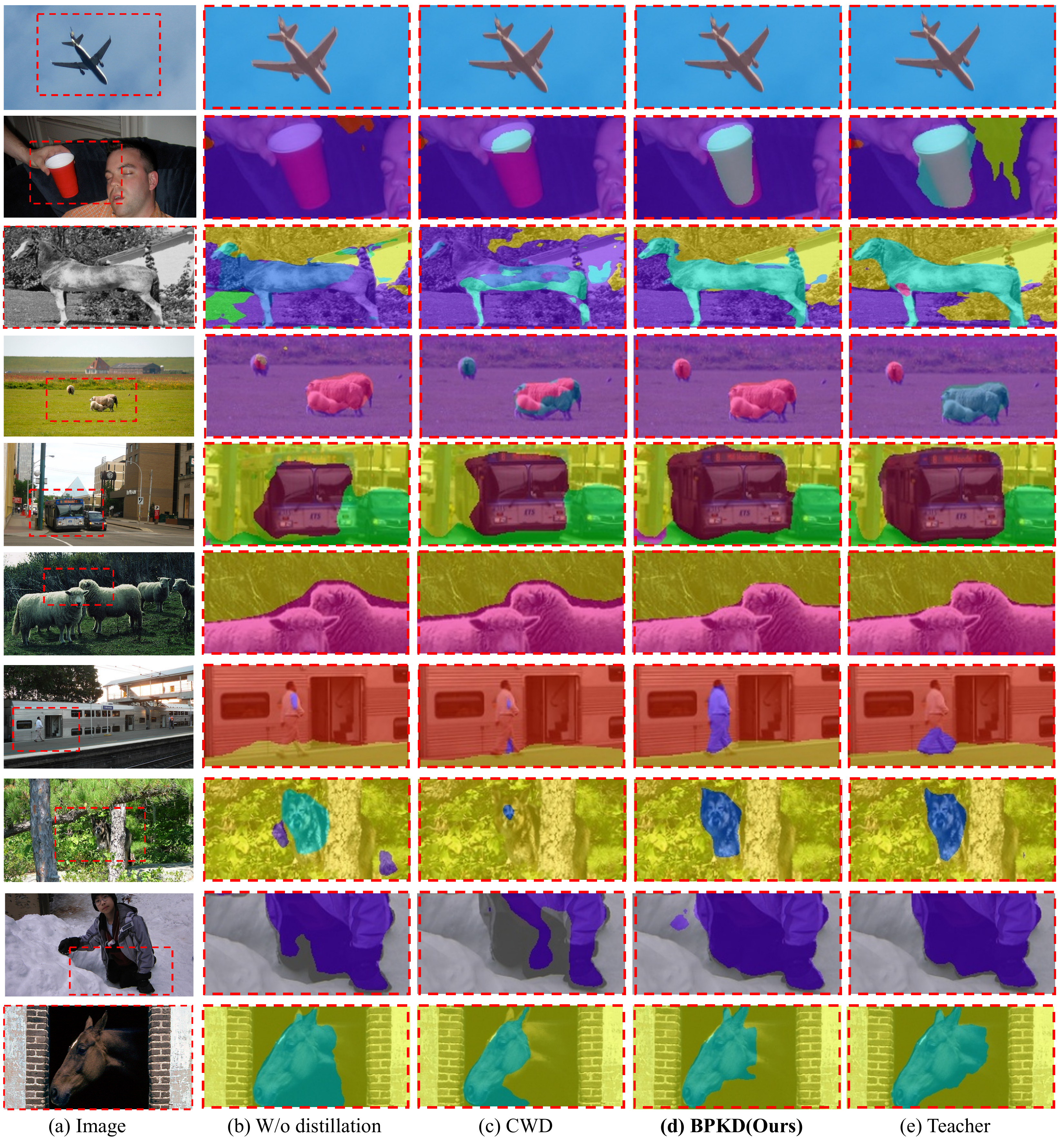}}
   \caption{
	Qualitative results on the Pascal-Context \cite{mottaghi_cvpr14} validation set produced by PSPNet \cite{zhao2017pyramid} and ResNet18 network architecture: (a) Initial images, (b) w/o distillation scheme, (c) sate-of-the-art method channel-wise distillation \cite{shu2021channel} , (d) \textbf{BPKD} our method, (e) teacher. This figure shows that our methods segment the small complex objects with explicit boundaries.  Zoom in for a better view.
 In the first row, BPKD exhibits more fine-grained edge segmentation compared to the non-distilled student network. In the second row, neither the student network nor CWD manages to achieve effective segmentation of the cup, owing to complex angles and environmental factors. Our approach, however, accomplishes accurate segmentation by leveraging prior knowledge of the cup's boundary and supervised learning on the main body of the cup. In the remaining examples, BPKD demonstrates more intricate and advanced segmentation results, thereby revealing that under the influence of edge loss, the entire network's convergence space benefits from shape and spatial priors across different classes. This, in turn, implicitly grants better main-body supervision and assistance, ultimately achieving state-of-the-art (SOTA) distillation results in segmentation.
	}
   \label{fig:qualitative_ade}
   \vspace{-1em}
\end{figure*}

\end{appendices}

\end{document}